# Artificial Intelligence and Statistical Techniques in Short-Term Load Forecasting: A Review

Ali Bou Nassif[1], Bassel Soudan[2], Mohammad Azzeh[3], Imtinan Attilli[4], Omar AlMulla[5]

**Abstract** – *Electrical utilities depend on short-term demand forecasting to proactively adjust production and distribution in anticipation of major variations. This systematic review analyzes 240 works published in scholarly journals between 2000 and 2019 that focus on applying Artificial Intelligence (AI), statistical, and hybrid models to short-term load forecasting (STLF). This work represents the most comprehensive review of works on this subject to date. A complete analysis of the literature is conducted to identify the most popular and accurate techniques as well as existing gaps. The findings show that although Artificial Neural Networks (ANN) continue to be the most commonly used standalone technique, researchers have been exceedingly opting for hybrid combinations of different techniques to leverage the combined advantages of individual methods. The review demonstrates that it is commonly possible with these hybrid combinations to achieve prediction accuracy exceeding 99%. The most successful duration for short-term forecasting has been identified as prediction for a duration of one day at an hourly interval. The review has identified a deficiency in access to datasets needed for training of the models. A significant gap has been identified in researching regions other than Asia, Europe, North America, and Australia.*

## I. Introduction

Technological advancements achieved during the past few decades have essentially become basic necessities of civilization as it enters the third millennium. These advancements share an almost complete dependence on a stable and continuous supply of electricity. Achieving this stability and continuity requires careful and continuous planning by the utilities. This can only be achieved through forecasting demand expectations and planning production in a proactive manner.

Utilities and governments attempt load forecasting for different time periods. Long-term load forecasting (LTLF) is used to predict expected demand growth for several years forward so that utilities and governments can plan service expansion. LTLF depends mostly on predictions of large-scale demographic and urban development variations. Medium-term load forecasting (MTLF) is typically carried out for future periods up to a few years in order to enact fuel supply contracts or schedule major maintenance activities [1]. MTLF depends on predictions of weather variations across the year and may integrate major load-affecting events (such as an exposition, or a locally-hosted global activity) into its prediction models.

Short-term load forecasting (STLF) on the other hand has a very short forward projection time frame and is typically intended for planning responses to predicted imminent variations in supply and/or demand. The timeframe for STLF is usually in the range of hours to possibly a small number of days forward. The target of STLF is typically the desire to maintain supply quality and possibly attain cost reduction by anticipating weather and demand variations. This is especially true if renewable sources are included in the system's generation mix. STLF is also desirable for possible load-shifting and load-scheduling applications that mitigate extreme demand peaks. Additionally, utilities depend on STLF to proactively plan bringing additional production online for sources that require time to spin-up before producing output (such as generators). Also, in a multi-generation system, STLF can be used to plan the dispatch schedule for the different sources [2].

Timeliness and accuracy in STLF is essential to maintaining output quality. If the forecasting model under-predicts the demand, the system will possibly need to utilize expensive quick-response backup mechanisms to compensate for the supply shortfall. On the other hand, if the model over-predicts, there will be surplus generation requiring the use of expensive storage or possibly electricity dumping.

STLF assimilates knowledge of current and historical conditions to predict future trends. Numerous possible inputs need to be taken into consideration to improve the forecast accuracy. Possible inputs include historical demand trends, expected weather patterns, time of the day, as well as the type of day (workday or a holiday). Classical modeling techniques are not able to account for all the different factors affecting the forecast. Therefore, the utilization of Artificial Intelligence (AI) techniques has become commonplace. Practitioners and researchers have investigated a myriad of models and techniques for STLF with the aim of attaining best possible prediction accuracy. There has been a significant variety of techniques used in the field. It is therefore worth conducting a review to determine the range of techniques utilized, their performance in terms of STLF prediction accuracy, and what are the possible gaps that require attention.

The earliest review of STLF techniques was conducted in 2001 and was targeted exclusively at the use of Neural Network (NN) variations in STLF[3]. Another review was conducted in 2002 to survey *unconventional methods* used in STLF [4]. This review addressed the use of supervised learning, unsupervised learning, and self-organizing methods. Another review was published in 2003 that discussed the use of AI and statistical techniques in STLF in general terms [5]. It discussed the features that techniques have to follow (causality, reproducibility, functionality, sensitivity and simplicity). A limited review was conducted in 2006 that concentrated on comparing the accuracy of six specific univariate methods for STLF [6]. Another review published in 2007 concentrated on the use of variants of Autoregressive Integrated Moving Average (ARIMA) models for predicting the daily demand profiles [7]. While these reviews were useful in focusing attention on the field, they are quite dated by now.

More recent works have included a review published in 2012 that surveyed the use of hybrid models combining NNs with other techniques [8]. This review targeted load forecasting in general rather than focusing on STLF specifically. A more detailed review was also published in 2012 that discussed many forecasting models used in the literature until that point in time [9]. Another work was published in 2014 that concentrated on reviewing the use of Artificial Neural Network (ANN) and Support Vector Machine (SVM) methods in predicting electrical energy consumption [10]. This review also discussed how hybridization is used to find more accurate results than single techniques. More recently, a 2015 paper reviewed the use of ANN models for STLF [11]. The paper also discussed how ANN can be combined with stochastic learning techniques. The most recent review of the field was published in 2016 which discussed load forecasting techniques in general [12]. The work reviewed probabilistic electric load forecasting including notable techniques, methodologies and evaluation methods, and common misunderstandings. As will be shown later in this work, there has been more than 100 studies published on the use of prediction models for STLF since this last review was conducted. Therefore, it is worthwhile to conduct a more up-to-date evaluation of the field. It is also worthwhile to widen the scope to cover as many prediction methods as possible to get a full understanding of the work that has been carried out.

This work reviews the use of AI, statistical models, and hybrid models for STLF in the literature. The aim is to evaluate existing STLF techniques and identify possible gaps that can be targeted for future research. The review will survey recent works published between January 2000 and September 2019. Works earlier than 2000 are not included since STLF has developed significantly since that timeframe. It will also be shown that the amount of work published in the last few years far outweighs anything that was published previously in terms of quantity as well as variety. Given the systematic methodology and long coverage period, this work is aimed to be the most comprehensive review of techniques used for STLF in recent literature.

The following section explains the methodology used in this research, then Section III presents statistical evaluation of the review pool. Section IV discusses the different techniques identified in the literature, then section V answers the research questions and discusses review findings. Section VI identifies the limitations of this research. Finally, Section VII presents concluding remarks and proposes future work. Significant details of all the research works included in the review pool are tabulated in the appendix B.

## II. Methodology

The methodology proposed by Kitchenham and Charters was selected for this review to ensure impartiality [13]. It also ensures repeatability by reducing search bias. The review protocol progresses through six main stages as summarized by the flowchart shown in Figure 1.

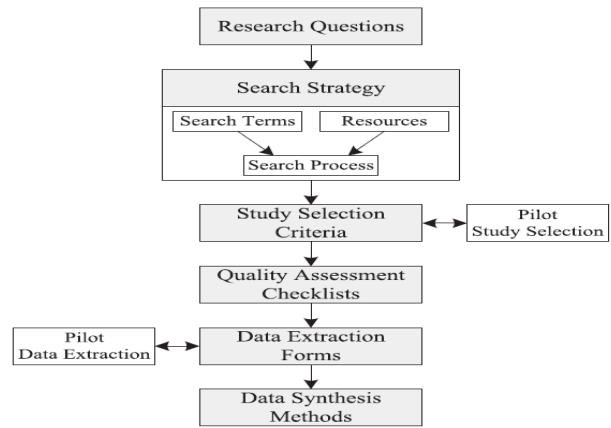

Figure 1. Stages of the review protocol

The main steps of the review protocol can be summarized as follows:

1. The objectives of the review were established as the identification and summarization of AI and statistical techniques used for STLF in current literature. The following research questions were devised to guide the analysis of the literature:
   RQ1: What are the techniques used for STLF in the literature?
   RQ2: Is the employed technique classified as Machine Learning, Non-machine Learning or Hybrids?
   RQ3: Is the employed model based on Regression or Classification?
   RQ4: Is the dataset used private or public and from which region?

RQ5: What do the authors imply by short-term?
RQ6: What are the parameters used in the forecasting?

2. Google scholar was selected as the main search database and concise search terms were devised to focus on the specific topic. The search focused on known primary publishers such as Elsevier, IEEE, Energies, Wiley, Taylor & Francis and Springer.

3. Each of the works identified in the previous stage was subjected to inclusion and exclusion criteria to determine the degree of match with the review topic. Initially, the Abstract, Key Words, and Conclusion sections were reviewed. However, detailed inspection was needed for many papers as their relevance was not clearly evident from these sections.

4. A pruning process was then used to keep the review pool manageable. Given the very significant number of high-quality journal publications included in the pool, a decision was taken to exclude conference publications. The justification is that journal publications are more detailed, have a wider scope, and typically pass through more stringent evaluation and review processes. At the end, the review pool comprised of 240 journal papers that met all the selection and relevance criteria. It is understood that excluding conference papers may omit some contributions in the field. However, it is the opinion of the authors that the journal paper-based review pool gives a highly adequate representation of the trends in general given its size and scope.

5. Each paper was evaluated individually to extract all possible relevant data. The process involved probing deeply into the content of many of the works in the review pool to answer the research questions posed earlier. A case-based reasoning decision tree methodology was also used to differentiate similarities.

6. It became evident that not all works discussed the techniques in a similar manner. Some works focused on more qualitative discussion rather than exact quantitative analysis. A data synthesis step became necessary to combine the different results and assimilate them into the framework of the review.

## III. Statistical Analysis of Review Pool

AT the end of the selection process discussed in the previous section, the review pool consisted of 240 journal papers. The following subsections present some statistics about these works to highlight the importance of this field, the amount of interest it is receiving, where to find relevant research work related to the field, and the popularity of the different prediction methods.

### III.1. Statistics Regarding Year of Publication

The chart in Figure 2 shows the distribution of the year of publication for the papers in the review pool. The trend line in the chart indicates consistently increasing publications per year reflective of continuous significant interest within the research community. The plot also shows a recent surge in publications consistent with the globally increased interest in application of AI techniques in different fields. It is to be noted that the collection of the review pool was carried out in the timeframe of September 2019, therefore it includes works published up to that point only.

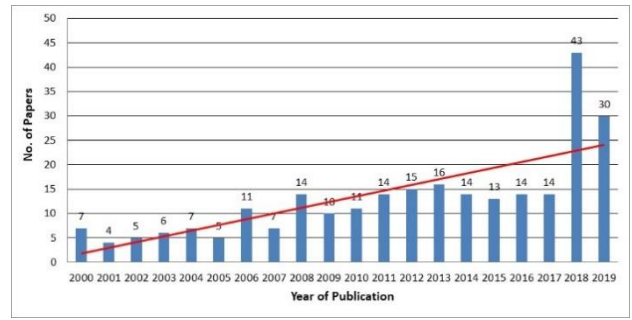

Figure 2. Distribution of the Review Pool by Year of Publication

### III.2. Statistics Regarding Publication Source

Research works related to the topic were extracted from 57 different journals. The diagram in Figure 3 identifies the top publication sources. The highest percentage of works (16%) were obtained from the journal Energy, followed by 12.5% from the IEEE Transactions on Power Systems, then 9% from Electric Power Systems Research. Other top sources were: Electrical Power and Energy Systems, Applied Energy, Energies, Expert Systems with Applications, and Applied Soft Computing. Other sources with fewer contributions were: Energy and Buildings, Energy Policy, International Journal of Forecasting, Energy Conversion and Management, and Neurocomputing. The journals Computer and Industrial Engineering, Fuzzy Sets and Systems, IEEE Access, IEEE Transactions on Industrial Electronics and Knowledge-Based Systems contributed only 3 papers each.

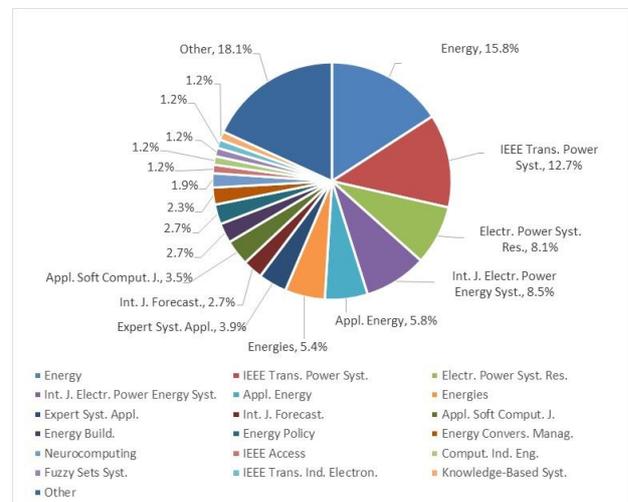

Figure 3. Publication Sources for Works in the Review Pool

### III.3. Forecasting Techniques Used in STLF

The works in the review pool were categorized into three broad groups based on the type of forecasting technique: standalone Machine Learning (ML) techniques, standalone Non-Machine Learning techniques and Hybrid Techniques (techniques employing multiple models). The chart in Figure 4 shows the distribution of the works based on this categorization. Standalone ML techniques accounted for 43% of works while standalone non-machine learning techniques accounted for only 13% of the works. On the other hand, about 44% of works reported using a hybrid model. This gives an indication how the popularity of hybrid techniques is now starting to surpass that of standalone ML-based techniques.

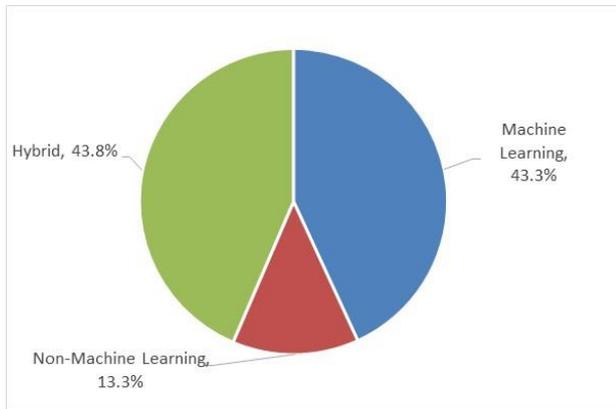

Figure 4. Category of Forecasting Techniques used in the Review Pool

Analysis of the review pool revealed that researchers employed a significant variety of forecasting techniques. The following is a listing of the techniques identified in the review pool based on the categorization mentioned earlier. The techniques are listed within each category in order of popularity.

- Standalone AI Models
  1. ANN
  2. Time Series Models
  3. Fuzzy Logic
  4. Particle Swarm Optimization (PSO)
  5. ARIMA
  6. Decomposition Models
  7. Self-Organizing Map Model (SOM)
  8. Ant Colony Optimization (ACO)
  9. Grey Prediction
  10. Genetic Algorithm (GA)
  11. Kalman Filtering algorithm

- Standalone Statistical Models
  1. Regression models
  2. Support Vector Regression (SVR)
  3. Bayesian Vector Auto-Regression (BVAR)

- Hybrid Models

The chart in Figure 5 shows that ANN was the most popular as it was used standalone in 20% of the works, and as a hybridized combination in another 20%.

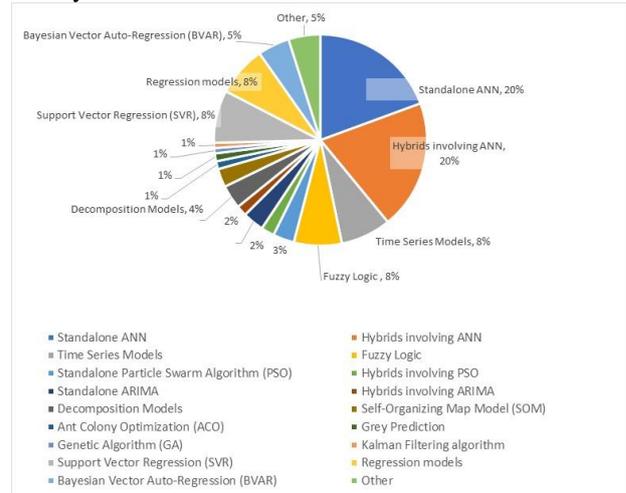

Figure 5. Distribution of STLF Techniques in the Review Pool

### III.4. Regression vs Classification in STLF

The chart in Figure 6 shows that while researchers used regression techniques significantly more frequently (87% of the works), still some 13% of the works utilized classification techniques.

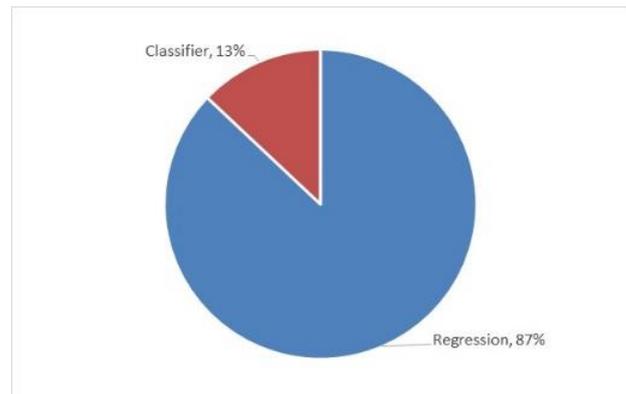

Figure 6. Regressions vs Classification Techniques in the Review Pool

### III.5. Performance Evaluation Criteria

Several performance evaluation criteria have been used in the literature to evaluate the accuracy of the predictive models. The chart in Figure 7 shows the prevalence of the different criteria in the review pool. The most common criteria were (in order of prevalence):
1. Mean Absolute Percentage Error – MAPE
2. Mean Squared Error – MSE
3. Root Mean Squared Error – RMSE
4. Mean Absolute Error – MAE
5. Average Percentage Error – APE

The percentages in the chart in Figure 7 add up to more than 100% because a number of studies utilized a number of evaluation criteria at the same time. Such works were counted in the statistics of each of the utilized criterion. Finally, about 8% of the works did not identify the evaluation criteria.

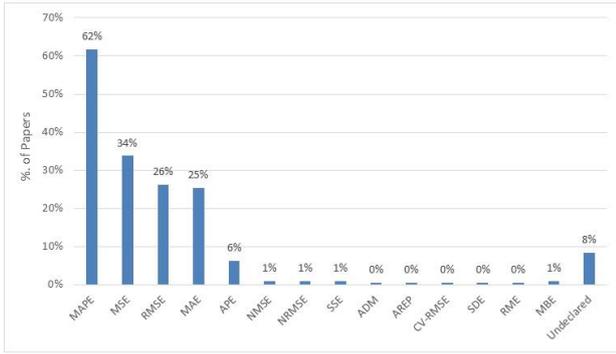

Figure 7. Performance Evaluation Criteria Utilized in the Review Pool

### III.6. Forecasting Interval and Duration

It was noted that researchers used different intervals to reflect '*short-term*' and implemented forecasting for different durations. The chart in Figure 8 shows the variation in the forecasting interval. The most common forecasting interval was to forecast at an hourly rate (82% of the works). It is to be noted that some works opted to perform forecasting for more than one interval. As an example, there were several works that forecasted both on an hourly and daily bases. These cases were reflected in the statistics of each of the forecasted intervals.

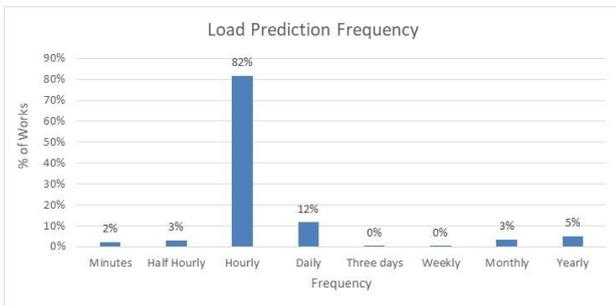

Figure 8. Forecasting Intervals Utilized in the Review Pool

Another variation between the works in the review pool was in the forecasting duration (as shown in Figure 9). The most common forecasting duration (30%) was for one day at a time. Less than 20% of the publications carried the forecasting for a duration of a full week. A small number of works carried out the forecasting for extended periods such as a full month or a full year. About half of the works in the review pool (49%) did not mention the duration.

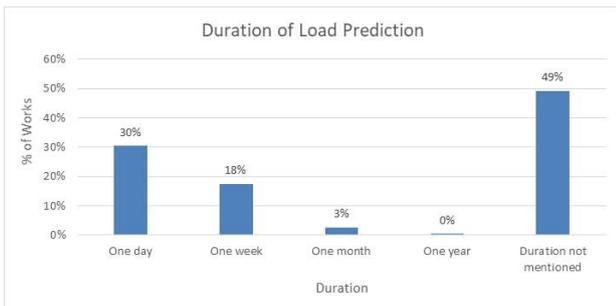

Figure 9. Forecasting Durations

The statistics presented in Figure 8 and Figure 9 indicate that the most commonly agreed upon definition of *short-term* among researchers of STLF is prediction for a duration of one day at an hourly interval.

### III.7. Dataset Privacy and Region of Study

It is very important to rely on historical datasets in predicting future load. Yet, it is often difficult to obtain detailed data from utilities. Even when utilities make the load datasets available, they do so with limited details. This often leads researchers to depend on private datasets. Private datasets are defined as those that are accessible only to the authors of the work. The chart in Figure 10 shows the statistics on reported accessibility of datasets used in the review pool. 61% of the works reported using private datasets. Another 3% of works did not clarify whether the dataset is private or publicly accessible.

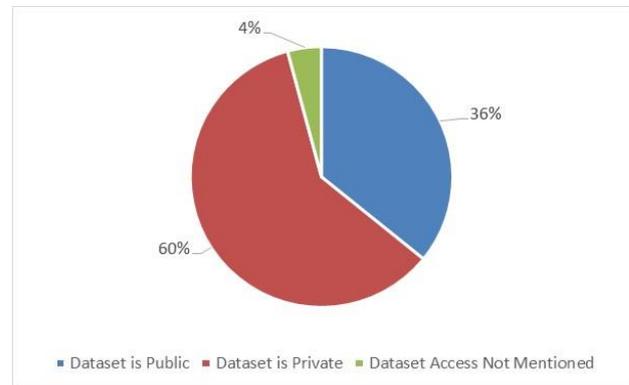

Figure 10. Accessibility of Dataset

The chart in Figure 11 shows that 36% of the works in the review pool utilized datasets from Asia making it the most often analyzed region. It was followed by Europe with 29%, then North America with 18%, and Australia with 11%. It is worth noting that South Africa, South America, and the Middle East appeared in the fewest number of works on STLF.

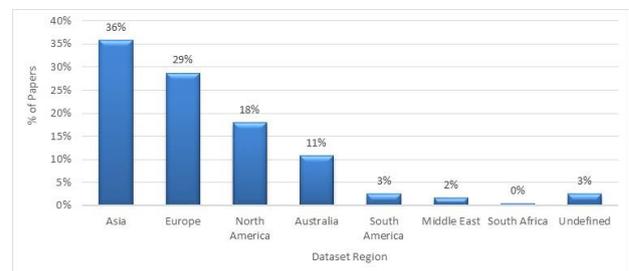

Figure 11. Regional Distribution of Datasets

## IV. Techniques used for STLF

This section discusses the utilization of the different forecasting techniques in the review pool following the categorization mentioned earlier.

*IV.1.  Standalone ML Techniques*

**Artificial Neural Networks (ANN)**

ANN was extensively used in the past for electrical load forecasting. However, its usage has decreased as many AI techniques are being developed. Nonetheless, standalone ANN implementations represented the most prevalent (non-hybrid) technique with 20% of the works in the review pool. In general, researchers reported that NN provides high-precision, and requires fewer samples and interpolation points.

In practice, utilities typically use ANN in their load forecasting departments since it is a relatively well-understood and proven technique [14]–[20]. In research, Carpinteiro et al. used ANN to predict the electric load once every hour for the span of the next 24 hours as required by Brazil Utility [21]. Improved results were obtained when a cascaded NN with cross-validation was applied to data from the Pennsylvania, Jersey, Maryland (PJM) and New York electricity markets [14]. A number of research works have examined the application of Wavelet Analysis or Wavelet Transform (a form of ANN) to STLF since it is known for its simplicity and easy combination with other forecasting models [8], [15], [22]–[28]. Researchers also examined the use of Radial Basis Functions (RBF) to predict the load in New England [29].

Some of the more recent works attempted the use of recently introduced ANN advancements and variations for the application of STLF. Their work proved that generalized neuron models have flexibility at both the aggregation and activation function level to cope with the non-linearity problems. Hernández et al. presented a two-stage prediction model based on ANN in order to estimate peak values of the demand curve [30]. Methaprayoon et al. showed that a multistage ANN model improves load forecast performance in both hour-ahead and day-ahead forecasting [31]. Furthermore, Chan et al. showed that a generalized learning strategy ANN can solve the problem of over-fitting caused by shortage of training data [32]. Siddharth Arora showed that Rule-Based Triple Seasonal Methods can improve the forecasting accuracy [33]. In addition, Abdel-Aal demonstrated that Abductive Network ML is a good tool for predicting the load [34]. Additionally, Hosseini and Gandomi showed that Gene Expression Programming with ANN has the capability of predicting the peak load with high accuracy [35]. Researchers have shown that Bayesian Neural Networks (BNN) can derive uncertain intervals on the model output and therefore find an optimum solution because they are generally better in cross-validations [36] [37].

A number of works used relative humidity and ambient temperature as the only input parameters to the ANN model [1], [19], [46], [47], [38]–[45]. This was justified based on the realization that temperature conditioning equipment account for the highest consumption of electrical energy. This approach was common in works targeting countries with unstable weather conditions.

The accuracy of ANN has been shown to improve when combined with other methods [48]–[53]. Panapakidis discussed how ANN can be clustered and characterized by high level of parameterization and efficiency. This model was tested on a set of buses covering urban, sub-urban and industrial loads [54]. Furthermore, Da-Silva applied Confidence Intervals based on NNs and found that it was very reliable since its intervals represent the true population for the sample [55]. Some works employed combinations of ANN methodology such as Entropy-Based Reduction (EBR) and Monte Carlo Algorithms [29][56]. A number of recent works proposed improving the accuracy using deep NN models [57]–[65].

**Time Series Models**

Time Series Modeling techniques build a model based on sequences of ordered numerical data points [66][67]. Safdarian et al. used mixed-integer programming to predict the next day load in Finland and results showed very close accurate predictions [66]. Additionally, Mastorocostas et al. applied orthogonal least-squares method for generating simple and efficient Takagi-Sugeno-Kang (TSK) fuzzy models [68]. It is a two-way technique that reduces complexity by discarding the unnecessary Fuzzy Basis Functions (FBF) and therefore results in accurate prediction [68]. Wang et al. applied non-linear fractal extrapolation algorithm expecting that it may improve the accuracy [69]. However, their results were comparable to other methods. Other time series methods used in the literature include Singular-Spectrum Analysis (SSA) which is known for its good ability in characterizing and predicting electric load [70]. The authors achieved reasonably effective prediction [70]. Researchers also reported accurate load forecasts using the multiple equation time approach [71][72].

A number of works compared time-series based models against ANN. Liu et al. used seasonal autoregressive integrated moving average with exogenous variables scheme and showed that it outperforms ANN [73]. Amjady compared the use of time series modeling against conventional techniques such as ANN [74]. Paparoditis and Sapatinas proposed a novel functional time-series methodology to predict the daily power load in Cyprus [75]. Their model accounted for weather variables such as humidity, wind speed and other factors and gave satisfactory predictions. Espinoza et al. used an individual method to predict the load demand of 245 substations based on Periodic Time Series and clustering problems [76]. With the availability of 40,000 data points for each substation STLF was estimated with good accuracy. De Grande et al. applied oriented load prediction model for enhancing prediction precision for large distributed systems [77]. Their results presented the best balancing efficiency gain for enhancing decision-making. García-Ascanio and Mate used interval time series combined with a vector auto-regressive model [78]. The latter model represented more accurate predictions compared to multi-layer perceptron. Details about the use of other time series models can be found in Table 1 in Appendix B [79]–[82].

**Fuzzy logic**

Researchers have used fuzzy logic methods for load forecasting because they are known for their low deviation

*A. B. Nassif, B. Soudan, M. Azzeh, I. Attilli, O. AlMulla*

between predicted and actual electric load. Accordingly, there has been a number of works published about using fuzzy logic as a standalone method for STLF [83][84]. Other works proposed using fuzzy logic in conjunction with NNs to produce more accurate forecasts. Such a combination gives better results because NNs deal better with sensory data while fuzzy logic deals better with reasoning on a higher level [85][86][87]. It has been shown that fuzzy logic tend to converge faster than conventional methods [88][89].

Researchers have attempted to introduce certain variations to improve the speed and accuracy of fuzzy logic. Huang incorporated a data handling enhancement that reduces convergence time and improves accuracy by identifying significant inputs to include in the prediction process [90]. Yang et al. used fuzzy logic based on chaotic dynamics reconstruction in order to reduce the computation error influences on correlation dimension estimation [91]. Moreover, fuzzy regression and a combination of fuzzy logic and knowledge based systems have been used to reduce the volatile property of temperature which produced better efficiency and a decrease in the absolute error [92][93][94].

### Autoregressive Integrated Moving Average (ARIMA)

ARIMA models are considered a stable, reliable, and effective method for forecasting electrical demand as they focus on the evolution of the intraday cycle [95][96]. Some researchers showed that it is possible to achieve good prediction accuracy using ARIMA as a standalone approach [97]. Others attempted improving the precision by incorporating residual modification models [98]. Lee and Ko designed a model using lifting scheme with ARIMA and tested it on five cases with different load datasets [99]. Their results showed higher forecasting accuracy compared to traditional ARIMA. A univariate prediction ARIMA model was also applied to achieve exponential smoothing [100]. Likewise, Taylor used the Double Seasonal Exponential Smoothing ARIMA model to enable smooth switching between demand forecasting methods [101]. Researches have also investigated the use of SARIMAX (Seasonal ARIMA with exogenous variables) technique for STLF prediction [102], [103].

### Decomposition Models

Decomposition modeling is a statistical method based on time series [104]. This method seeks to construct deterministic and non-deterministic components from a dataset in order to analyze them and forecast future behavior. Almeshaiei and Soltan proposed a methodology based on decomposition and segmentation to construct a load forecasting model [105]. Similarly, Taylor applied exponentially weighted methods based on singular value decomposition [106]. This method was shown to achieve simpler and more efficient model formulation. Høverstad et al. used evolution for parameter tuning with seasonal decomposition [107]. They showed that this method improves the load estimation. Zhang et al. developed a decomposition-ensemble model that integrates SSA, an SVM, ARIMA and the cuckoo search algorithm [108].

### Grey Prediction

This model combines a partial theoretical structure with date for predicting future data. Dong Li et al. used an improved grey dynamic model which gave more accurate forecasting taking into consideration the effects of weather conditions, economy changes and society information [109]. Jin et al. also applied grey correlation contest modeling in order to cover the impact of different climatic and societal factors [110]. They used it to determine the best strategy for forecasting smooth predicted load. Researchers also used an improved grey prediction model which combines data transformation and combination interpolation optimization [111]. Wang et al. developed single-linear, hybrid-linear, and non-linear forecasting techniques based on grey theory to improve the accuracy of STLF for China and India [112].

### Kalman Filtering algorithm

Kalman filter is a well-known linear quadratic estimator that uses series measurements of old data to produce estimated unknown variables more accurately compared to methods utilizing single measurements. Usually, Kalman filters use Bayesian inference with a joint probability distribution over variables of each time frame. Guan et al. used this method to capture different features of load components in forecasting [113]. Their results showed only a small deviation between the predicted and real values. Al-Hamadi and Soliman reported less than 1% error using Kalman filtering with a moving window weather and load model [114].

### Particle Swarm Optimization (PSO)

PSO presents the ability to converge towards a global minimum on an error surface, thus minimizing the error in forecasted load. Huang et al. concluded that this method could be used to improve the ARIMA with exogenous variable model [115]. EL-Telbany and El-Karmi compared the performance of PSO to Back-Propagation (BP) and Autoregressive-Moving Average (ARMA) for predicting electrical load in Jordan [116]. Although PSO learning and its search algorithm required more function evaluations, it performed better than the BP algorithm.

### Self-Organizing Map Model (SOM)

SOM is a type of ANN that produces a low dimensional and discretized representation of input space of the training samples. This method applies competitive learning using neighborhood function to preserve properties of the input space. Carpinteiro and Da Silva used an SOM model for predicting hourly load for a Brazilian utility [117]. SOM was also used for forecasting the Spanish electricity market [118]. Mao et al. proposed using Self Organizing Fuzzy NNs with bilevel optimization to simplify the forecasting task [119]. These studies showed good results for large forecasting horizons.

### Ant Colony Optimization (ACO)

Using ACO, researchers were able to forecast net electricity energy generation for Turkey with a 5% maximum prediction error [120].

*Genetic Algorithm (GA)*

Only one work has been found in the review pool where GA was used standalone to perform STLF [121]. GA was mainly used for tuning the parameters.

*IV.2. Standalone Statistical Models*

*Regression Models*

Regression modeling is a statistical process in predicting relationships in a multivariate problem. It examines the effect of change in an independent variable on the value of a dependent variable [122]. In general, regression models produce accurate predictions of the future of variable behaviors based on historical data [123][124][125]. Because of its ability to predict future behavior, regression has been applied successfully in the field of STLF [122][126]. Examples include Zamo et al. who used regression methods for forecasting solar electricity production based on random forecasts for photovoltaic output [127]. Maia and Gonçalves applied special adaptive recursive digital filtering as a regression model [128]. Similarly, Amaral et al. applied a smooth transition periodic autoregressive model and evaluated its performance against alternative load forecasting models [129]. The method showed more accurate predictions compared to ANN models. Researchers used linear autoregressive models to predict electrical load in Brazil and Serbia and obtained better results than ANN models [130][131]. Fuzzy Linear Regression Method was also used by Song et al. to predict holiday load in Korea [132]. They were able to achieve a maximum prediction error of 3.57% compared to actual.

Dudek used linear regression modeling to tackle the problem of filtering the affecting variables in seasonal and trend variations of longer periods than daily variables [133]. Yang and Stenzel applied a novel regression tree incrimination method on historical data to predict future load. Their work improved prediction accuracy by eliminating interference caused by deserted boarder points [134]. Fan and Hyndman used a semi-parametric additive model that produced very accurate load predictions which reduced power over generation [135]. Additionally, there has been a number of recent works that have also investigated the use of regression models [136]–[144]. Table 2 in Appendix B shows the comparative accuracy of the different regression models used in the literature.

*Support Vector Regression (SVR)*

SVR models are usually applied to structural risk minimization to reduce the generalization error upper bound instead of minimizing training error as ANN models do [145][146]. Hong applied this method and found that SVR with immune algorithm produced more accurate results than both regression and ANN models [146][147]. Furthermore, Wang et al. combined an SVR model with a seasonal adjustment model to help in producing smooth data series and effectively remove seasonal variations [148]. The procedure's performance was satisfactory based on analysis of variance with validation and relative verification. Likewise, Niu et al. merged an SVR model with an ACO algorithm to process larger amounts of load data. The combination reduced redundant information and found optimal feature subsets in a fuzzy-rough process of data reduction [149]. The results showed more stable and accurate prediction for load forecasting than ARMA and ANN models. Moreover, Ren et al. used a random vector functional link network in order to avoid mapping weighted inputs at saturation region [150]. They compared the performance of this method with ARIMA and ANN models. Their results showed better hourly forecasting performance while ARIMA degraded performance on seasonal time series. Researchers have proposed the use of SVR for implementing a hybrid filter–wrapper approach for feature selection incorporating a Partial Mutual Information (PMI) based filter and the firefly algorithm as the wrapper [151]. They claim superiority to other feature selection methods. Li et al. investigated using a Subsampled Support Vector Regression Ensemble (SSVRE) to improve the computational accuracy and efficiency [152].

*Bayesian Vector Auto-Regression (BVAR)*

BVAR is a technique known for its significant advantages over classical NN methods where it can cite automatic tuning of regularization coefficients. A number of works investigated the use of different regression models (BVAR, SVR, and inverse modeling) for predicting electricity demand [153]–[155]. In general, BVAR performed slightly worse than SVR and other regression models.

*IV.3. Hybrid Models*

The above discussion outlined the use of standalone techniques in the literature for STLF forecasting. A significant amount of work has also been done to investigate possible improvements that can be attained using hybrid forecasting models. Hybrid models are combinations of two or more techniques that attempt to leverage the points of strength of the different models to improve the accuracy of the prediction. Such hybrid models have been promoted in the literature as the future of AI due to their performance improvements.

A number of hybrid combinations of different NN implementations have been proosed [156], [157], [166], [158]–[165]. Dong-xiao Niu used BNNs combined with a hybrid Monte Carlo algorithm to improve the forecasting performance [156]. Chaturvedi et al. used improved generalized neuron model for solving non-linear complex problems that relatively require larger number of hidden nodes [157]. Some of the hybrid methods investigated improving the prediction accuracy by combining NN implementations with other AI Techniques [167], [168], [177]–[180], [169]–[176]. Other works investigated the different combinations of AI techniques with the Wavelet transform [181], [182], [191]–[195], [183]–[190].

A number of works combined PSO with other AI techniques [196]–[199]. Selakov et al. used PSO in conjunction with SVM for STLF subject to significant temperature variations [196]. Their results showed better accuracy compared to classical methods on recent history and similar days. A number of researchers proposed combining SOM with other AI techniques to improve the efficiency and accuracy [200]–[203]. An example includes the use of SOM as a clustering method with RBF as the classification method for the prediction [200].

In general, most of the hybrid implementations were based on combinations of different AI techniques [204], [205], [214]–[223], [206], [224]–[230], [207]–[213]. As an example, Shakouri et al. mixed fuzzy regression with fuzzy rule-based Takagi–Sugeno–Kang (TSK) model to study the impact of weather changes on electrical load consumption [204]. Ghanbari et al. combined ACO and GA to simulate fluctuations of electrical demand [205]. They reported more stable results and accurate load forecasting compared to ANN. Niu et al. evaluated a nonlinear Multilayer Perceptron (MLP) optimized using an adaptive PSO [206].

Researchers also investigated the combination of AI techniques and various regression models. For example, researchers proposed the combination of a regression model with different NN implementations [231]–[239]. Researchers also proposed combination of different AI techniques with SVR and BVAR [240]–[246]. In particular, some works evaluated the combination of Empirical Mode Decomposition (EMD) with other AI and regression models [247]–[252]. Finally, some of the work on hybrid models investigated different combinations of regression models [253]–[260].

Altogether, almost 44% of the works in the review pool involve some form of a hybrid model as indicated by Table 3 in Appendix B. Actually, a significant portion of the most recent work targeted hybrid models exclusively [261], [262], [271], [272], [263]–[270].

## V. Review Findings

A number of research questions were raised in section **Error! Reference source not found.** to guide the review. The following are the answers to these research questions based on the review:

**RQ1: What are the techniques used for STLF in the literature?**

The most common techniques used in the literature for STLF are (ordered from the most common to the least): ANN, Hybrids combining ANN with other techniques, Regression Models, SVR, Time Series Models, Fuzzy Logic, PSO, BVAR, ARIMA, Decomposition Models, Kalman Filtering, SOM, Grey Prediction, ACO, and GA. It is noteworthy that a good amount of research has been done using hybrid combinations of these techniques.

**RQ2: Is the employed technique classified as Machine Learning, Non-machine Learning or Hybrids?**

ML techniques were used in about 43% of the works, Hybrid techniques were used in about 44%, and Non-ML were used in about 13%.

**RQ3: Is the employed model based on Regression or Classification?**

Regression techniques were used in 87% of the works while only 13% applied classification techniques.

**RQ4: Is the dataset used private or public and from which region?**

Private datasets were used in 61% of the works while 36% of the researchers used publicly accessible datasets. It was not possible to identify the dataset for 3% of the works. Datasets from Asia were the most commonly analyzed followed by Europe, North America, and Australia. Very few works targeted data from South Africa, South America, and the Middle East.

**RQ5: What do the authors imply by short-term?**

There was a large variation in the understanding of '*short-term*'. However, the statistics presented in Figure 8 and Figure 9 indicate that the most common understanding of *short-term* was prediction for a duration of one day at an hourly interval.

**RQ6: What are the parameters used in the forecasting?**

The most common factors used to guide the STLF prediction were: weather variables (temperature, humidity, etc…), population growth, and type of day (weekdays, weekends, etc…).

The following findings can be extracted based on comparison of reported results in the review pool (presented in Table 1 through Table 3 in Appendix B):

- It is possible to achieve very high prediction accuracy (reported MAPE < 1%) for STLF. Actually, 55 of the works in the review pool reported a prediction error below 1%. A similar number reported prediction error between 1% and 2%. Therefore, any future work on STLF should target less than 1% prediction error to be competitive with existing works in the literature.

- Most standalone ANN implementations achieved a reported prediction error between 1% and 2%. While there are a number of works that reported higher accuracy (MAPE < 1%), these works are very few and seem to be the outliers. On the other hand, a significant number of works utilizing standalone ANN produced errors exceeding 2%. It is clear that other techniques (especially hybrids) produce much better results for STLF than standalone NN.

- Overall, standalone time series based applications reported higher prediction accuracy with numerous works reporting error rates below 1%. In general, time series models presented the highest accuracy amongst all standalone machine learning models. This can be attributed to the fact that time series models agree better with the principle of STLF since in the latter, the values of power load records are saved in sequential order.

- All standalone fuzzy logic implementations produced relatively high error rates (MAPE approaching 3%).

- Even when combined with other methods in a hybrid implementation, their accuracy was still lower than other prediction methods.

- Standalone ARIMA models seem to fair reasonably well in terms of accuracy with about half the implementations reporting error rates below 1%. It is to be noted that standalone ARIMA implementations fared better than hybrid ARIMA combinations, with most hybrids producing prediction errors approaching 3%.

- There were infrequent implementations using other ML techniques to create a statistically significant case. Still, in general, most of these implementations produced prediction errors approaching and exceeding 2%.

- In general, standalone statistical models were relatively very poor as compared to the other methods presented in the literature. The majority of works reported significantly high prediction errors and only a handful of implementations were able to achieve an error below 1%. Hybrids combining regression models with ML implementations performed better overall compared to standalone regression models.

- While hybrid models were expected to outperform standalone techniques, the results show mixed results in terms of prediction accuracy. About equal numbers of works based on hybrid models reported prediction error below 1%, between 1% and 2%, and higher than 2%. Combinations incorporating hybrid implementations of NN produced improved prediction errors compared to standalone NN. Also, Wavelet transform combined with other techniques produced predictions with better accuracy (most of the works reported errors below 1%). On the other hand, combining NN with other ML techniques produced significant errors (approaching and exceeding 2%). Lastly, incorporating regression models with ML techniques did not produce any significant improvement in the prediction accuracy. These results seem to suggest that hybrid models did not contribute to improving the accuracy. On the other hand, hybrid models have a significant disadvantage in terms of higher computational time compared to standalone models.

## VI. Limitations of this Review

This review focuses mainly on AI and statistical techniques applied in STLF. Various techniques were discussed, and different results were reported. Although many of the researchers utilized similar techniques, their implementation approaches were different and produced different results. Therefore, it was not possible to readily produce comparative results.

A significant portion of the studies used datasets that are not publicly accessible. This casts a flaw on the drawn conclusions since other researchers would not be able to replicate the studies and validate or generalize the results. The works targeted in this review were journal papers published after the year 2000. This criterion was chosen because we believe that researchers are interested in the most recent research directions in the field. A quick search for works published before 2000 resulted in only a few publications. We believe that omitting these works did not significantly affect the results of the review as presented. We understand that such papers might still be useful, but we prefer to focus on the more recent research directions.

Some of the studies in the review pool did not report accuracy indicators for their prediction techniques, or they used uncommon accuracy measures. These studies have been excluded from accuracy analysis even though they might be still useful.

Although the topic of Short-term power load prediction is defined as a regression problem where the output is numeric, some papers applied classification models where the output is presented as a label. This limited the ability of comparing their results with works employing regression-based models.

Since the search identified an enormous number of works related to the topic, this review excluded all conference papers and focused only on works published in scientific journals. This was justified based on the premise that conference papers are typically significantly limited by the number of pages and therefore many details have to be omitted. Journal papers on the other hand do not have the same level of length restriction and pass through a more stringent review and validation process. Moreover, it was noted during the examination process that several papers in the review pool were conference papers that had been extended to journal papers. Excluding conference papers was necessary to keep the review manageable but may have resulted in excluding some papers that would have contributed to the results.

## VII. Conclusions and Future Work

This systematic review surveyed the applications of AI and Statistical techniques in STLF in recent research publications. The review covered published work between 2000 and 2019 inclusive. A large number of publications were first identified which made it necessary to concentrate only on journal publications since they tend to have a wider scope, and typically pass through more stringent evaluation and review processes. The review pool at the end comprised of 240 works from five primary libraries of publications. The number and trend of publications shows significant continued interest in the research community.

The review process categorized the publications based on the type of prediction algorithm. ML algorithms accounted for about 43%, Non-machine learning techniques accounted for only about 13%, with the remaining 44% being hybrid techniques that combine two or more prediction models. While ANN remains the most popular standalone technique (20% of the works in the review pool), almost half of the publications opted for hybrid algorithms to improve the prediction accuracy. Even though STLF lends itself more to regression-based technique, about 13% of the works performed the prediction using classification-based methods reasonably successfully. Researchers used a variety of performance evaluation criteria. The most common by far was MAPE followed by MSE, RMSE, and MAE.

Access to historical data seems to be a general issue. Most of the publications (61%) were based on private datasets and only about one third were based on publicly accessible data. This poses a reservation on the reproducibility of the results. Research also seems to be targeted towards the more developed regions in the world. Close to 95% of the works were based on datasets from Asia, Europe, North America, and Australia (in that order). This may indicate either lack of data from the other regions, or lack of research interest. Researchers have opted for different prediction frequencies ranging from minute-by-minute to a full year at a time. Still, there is general agreement on using hourly predictions (82% of the works) as the appropriate frequency for STLF.

The reported results show that it is possible to achieve very high accuracy with prediction errors below 1%. Any method that achieves a prediction error higher than 2% would be lacking compared to the state-of-the-art reported in the literature. Standalone NN are no longer the best prediction methods for STLF, with numerous other techniques achieving much better prediction accuracy. Time series models and hybrids incorporating Wavelet transform seem to be currently the best performing methods. While hybrids were espoused to improve the prediction accuracy, the results in the review pool do not support this hypothesis. Contrarily, hybrids require significantly increased execution time that is not justified by the current results.

Short-term load forecasting is an active field of research with strong interest in the community. There are clearly gaps were additional work can be targeted. Especially related to the use of new modern prediction techniques such as deep learning and big data classifications.

## VIII. Acknowledgement

The authors thank the University of Sharjah for its continuous support in research.

## References

[1] B. Satish, K. S. Swarup, S. Srinivas, and A. H. Rao, "Effect of temperature on short term load forecasting using an integrated ANN," *Electr. Power Syst. Res.*, vol. 72, no. 1, pp. 95–101, 2004.

[2] B. Soudan and A. Darya, "Autonomous smart switching control for off-grid hybrid PV/battery/diesel power system," *Energy*, vol. 211, Nov. 2020.

[3] H. S. Hippert, C. E. Pedreira, and R. C. Souza, "Neural networks for short-term load forecasting: A review and evaluation," *IEEE Trans. Power Syst.*, vol. 16, no. 1, pp. 44–55, 2001.

[4] D. Singh and S. P. Singh, "Self organization and learning methods in short term electric load forecasting: A review," *Electr. Power Components Syst.*, vol. 30, no. 10, pp. 1075–1089, 2002.

[5] K. Metaxiotis, A. Kagiannas, D. Askounis, and J. Psarras, "Artificial intelligence in short term electric load forecasting: A state-of-the-art survey for the researcher," *Energy Convers. Manag.*, vol. 44, no. 9, pp. 1525–1534, 2003.

[6] J. W. Taylor, L. M. de Menezes, and P. E. McSharry, "A comparison of univariate methods for forecasting electricity demand up to a day ahead," *Int. J. Forecast.*, vol. 22, no. 1, pp. 1–16, 2006.

[7] J. W. Taylor and P. E. McSharry, "Short-term load forecasting methods: An evaluation based on European data," *IEEE Trans. Power Syst.*, vol. 22, no. 4, pp. 2213–2219, 2007.

[8] M. Q. Raza and Z. Baharudin, "A review on short term load forecasting using hybrid neural network techniques," *PECon 2012 - 2012 IEEE Int. Conf. Power Energy*, no. December, pp. 846–851, 2012.

[9] L. Suganthi and A. A. Samuel, "Energy models for demand forecasting - A review," *Renew. Sustain. Energy Rev.*, vol. 16, no. 2, pp. 1223–1240, 2012.

[10] A. S. Ahmad *et al.*, "A review on applications of ANN and SVM for building electrical energy consumption forecasting," *Renew. Sustain. Energy Rev.*, vol. 33, pp. 102–109, 2014.

[11] A. Baliyan, K. Gaurav, and S. Kumar Mishra, "A review of short term load forecasting using artificial neural network models," *Procedia Comput. Sci.*, vol. 48, no. C, pp. 121–125, 2015.

[12] T. Hong and S. Fan, "Probabilistic electric load forecasting: A tutorial review," *Int. J. Forecast.*, vol. 32, no. 3, pp. 914–938, 2016.

[13] B. Kitchenham and S. Charters, "Guidelines for performing systematic literature reviews in software engineering," *Tech. report, Ver. 2.3 EBSE Tech. Report. EBSE*, 2007.

[14] S. Kouhi and F. Keynia, "A new cascade NN based method to short-term load forecast in deregulated electricity market," *Energy Convers. Manag.*, vol. 71, pp. 76–83, 2013.

[15] N. M. Pindoriya, S. N. Singh, and S. K. Singh, "An adaptive wavelet neural network-based energy price forecasting in electricity markets," *IEEE Trans. Power Syst.*, vol. 23, no. 3, pp. 1423–1432, 2008.

[16] P. J. Santos, A. G. Martins, and A. J. Pires, "Designing the input vector to ANN-based models for short-term load forecast in electricity distribution systems," *Int. J. Electr. Power Energy Syst.*, vol. 29, no. 4, pp. 338–347, 2007.

[17] D. Srinivasan, "Energy demand prediction using GMDH networks," *Neurocomputing*, vol. 72, no. 1–3, pp. 625–629, 2008.

[18] C. Hamzaçebi, "Forecasting of Turkey's net electricity energy consumption on sectoral bases," *Energy Policy*, vol. 35, no. 3, pp. 2009–2016, 2007.

[19] N. Ding, C. Benoit, G. Foggia, Y. Besanger, and F. Wurtz, "Neural network-based model design for short-term load forecast in distribution systems," *IEEE Trans. Power Syst.*, vol. 31, no. 1, pp. 72–81, 2016.

[20] J. P. S. Catalão, S. J. P. S. Mariano, V. M. F. Mendes, and L. A. F. M. Ferreira, "Short-term electricity prices forecasting in a competitive market: A neural network approach," *Electr. Power Syst. Res.*, vol. 77, no. 10, pp. 1297–1304, 2007.

[21] O. A. S. Carpinteiro, A. J. R. Reis, and A. P. A. Da Silva, "A hierarchical neural model in short-term load forecasting," *Appl. Soft Comput. J.*, vol. 4, no. 4, pp. 405–412, 2004.

[22] S. Kelo and S. Dudul, "A wavelet Elman neural network for short-term electrical load prediction under the influence of temperature," *Int. J. Electr. Power Energy Syst.*, vol. 43, no. 1, pp. 1063–1071, 2012.

[23] M. Y. Zhai, "A new method for short-term load forecasting based on fractal interpretation and wavelet analysis," *Int. J. Electr. Power Energy Syst.*, vol. 69, pp. 241–245, 2015.

[24] T. Nengling, J. Stenzel, and W. Hongxiao, "Techniques of applying wavelet transform into combined model for short-term load forecasting," *Electr. Power Syst. Res.*, vol. 76, no. 6–7, pp. 525–533, 2006.

[25] C. Guan, P. B. Luh, L. D. Michel, Y. Wang, and P. B. Friedland, "Very short-term load forecasting: Wavelet neural networks with data pre-filtering," *IEEE Trans. Power Syst.*, vol. 28, no. 1, pp. 30–41, 2013.

[26] G. T. Ribeiro, V. C. Mariani, and L. dos S. Coelho, "Enhanced ensemble structures using wavelet neural networks applied to

short-term load forecasting," *Eng. Appl. Artif. Intell.*, vol. 82, pp. 272–281, 2019.

[27] J. Wang, J. Wang, Y. Li, S. Zhu, and J. Zhao, "Techniques of applying wavelet de-noising into a combined model for short-term load forecasting," *Int. J. Electr. Power Energy Syst.*, vol. 62, pp. 816–824, 2014.

[28] Y. Chen *et al.*, "Short-term load forecasting: Similar day-based wavelet neural networks," *IEEE Trans. Power Syst.*, vol. 25, no. 1, pp. 322–330, 2010.

[29] C. Cecati, J. Kolbusz, P. Różycki, P. Siano, and B. M. Wilamowski, "A Novel RBF Training Algorithm for Short-Term Electric Load Forecasting and Comparative Studies," *IEEE Trans. Ind. Electron.*, vol. 62, no. 10, pp. 6519–6529, 2015.

[30] L. Hernández *et al.*, "Improved short-term load forecasting based on two-stage predictions with artificial neural networks in a microgrid environment," *Energies*, vol. 6, no. 9, pp. 4489–4507, 2013.

[31] K. Methaprayoon, W. J. Lee, S. Rasmiddatta, J. R. Liao, and R. J. Ross, "Multistage artificial neural network short-term load forecasting engine with front-end weather forecast," *IEEE Trans. Ind. Appl.*, vol. 43, no. 6, pp. 1410–1416, 2007.

[32] Z. S. H. Chan, H. W. Ngan, A. B. Rad, A. K. David, and N. Kasabov, "Short-term ANN load forecasting from limited data using generalization learning strategies," *Neurocomputing*, vol. 70, no. 1–3, pp. 409–419, 2006.

[33] S. Arora and J. W. Taylor, "Short-term forecasting of anomalous load using rule-based triple seasonal methods," *IEEE Trans. Power Syst.*, vol. 28, no. 3, pp. 3235–3242, 2013.

[34] R. E. Abdel-Aal, "Short-Term Hourly Load Forecasting Using Abductive Networks," *IEEE Trans. Power Syst.*, vol. 19, no. 1, pp. 164–173, 2004.

[35] S. S. S. Hosseini and A. H. Gandomi, "Short-term load forecasting of power systems by gene expression programming," *Neural Comput. Appl.*, vol. 21, no. 2, pp. 377–389, 2012.

[36] P. Lauret, E. Fock, R. N. Randrianarivony, and J. F. Manicom-Ramsamy, "Bayesian neural network approach to short time load forecasting," *Energy Convers. Manag.*, vol. 49, no. 5, pp. 1156–1166, 2008.

[37] H. S. Hippert and J. W. Taylor, "An evaluation of Bayesian techniques for controlling model complexity and selecting inputs in a neural network for short-term load forecasting," *Neural Networks*, vol. 23, no. 3, pp. 386–395, 2010.

[38] N. Kandil, R. Wamkeue, M. Saad, and S. Georges, "An efficient approach for short term load forecasting using artificial neural networks," *Int. J. Electr. Power Energy Syst.*, vol. 28, no. 8, pp. 525–530, 2006.

[39] A. Deihimi and H. Showkati, "Application of echo state networks in short-term electric load forecasting," *Energy*, vol. 39, no. 1, pp. 327–340, 2012.

[40] A. S. Khwaja, X. Zhang, A. Anpalagan, and B. Venkatesh, "Boosted neural networks for improved short-term electric load forecasting," *Electr. Power Syst. Res.*, vol. 143, pp. 431–437, 2017.

[41] K. Kalaitzakis, G. S. Stavrakakis, and E. M. Anagnostakis, "Short-term load forecasting based on artificial neural networks parallel implementation," *Electr. Power Syst. Res.*, vol. 63, no. 3, pp. 185–196, 2002.

[42] S. Fan, L. Chen, and W. J. Lee, "Short-term load forecasting using comprehensive combination based on multimeteorological information," *IEEE Trans. Ind. Appl.*, vol. 45, no. 4, pp. 1460–1466, 2009.

[43] W. Charytoniuk and M.-S. Chen, "Very Short-Term Load Forecasting Using Artificial Neural Networks," *IEEE Trans. Power Syst.*, vol. 15, no. 1, p. 263, 2000.

[44] M. Beccali, M. Cellura, V. Lo Brano, and A. Marvuglia, "Forecasting daily urban electric load profiles using artificial neural networks," *Energy Convers. Manag.*, vol. 45, no. 18–19, pp. 2879–2900, 2004.

[45] A. S. Khwaja, M. Naeem, A. Anpalagan, A. Venetsanopoulos, and B. Venkatesh, "Improved short-term load forecasting using bagged neural networks," *Electr. Power Syst. Res.*, vol. 125, pp. 109–115, 2015.

[46] E. González-Romera, M. Á. Jaramillo-Morán, and D. Carmona-Fernández, "Forecasting of the electric energy demand trend and monthly fluctuation with neural networks," *Comput. Ind. Eng.*, vol. 52, no. 3, pp. 336–343, 2007.

[47] N. Amjady, F. Keynia, and H. Zareipour, "Short-term load forecast of microgrids by a new bilevel prediction strategy," *IEEE Trans. Smart Grid*, vol. 1, no. 3, pp. 286–294, 2010.

[48] C. Kang, X. Cheng, Q. Xia, Y. Huang, and F. Gao, "Novel approach considering load-relative factors in short-term load forecasting," *Electr. Power Syst. Res.*, vol. 70, no. 2, pp. 99–107, 2004.

[49] A. Dolara, F. Grimaccia, S. Leva, M. Mussetta, and E. Ogliari, "A physical hybrid artificial neural network for short term forecasting of PV plant power output," *Energies*, vol. 8, no. 2, pp. 1138–1153, 2015.

[50] P. Li, Y. Li, Q. Xiong, Y. Chai, and Y. Zhang, "Application of a hybrid quantized Elman neural network in short-term load forecasting," *Int. J. Electr. Power Energy Syst.*, vol. 55, pp. 749–759, 2014.

[51] I. P. Panapakidis, "Application of hybrid computational intelligence models in short-term bus load forecasting," *Expert Syst. Appl.*, vol. 54, pp. 105–120, 2016.

[52] Z. Xiao, S. J. Ye, B. Zhong, and C. X. Sun, "BP neural network with rough set for short term load forecasting," *Expert Syst. Appl.*, vol. 36, no. 1, pp. 273–279, 2009.

[53] S. A. Villalba and C. Á. Bel, "Hybrid demand model for load estimation and short term load forecasting in distribution electric systems," *IEEE Trans. Power Deliv.*, vol. 15, no. 2, pp. 764–769, 2000.

[54] I. P. Panapakidis, "Clustering based day-ahead and hour-ahead bus load forecasting models," *Int. J. Electr. Power Energy Syst.*, vol. 80, pp. 171–178, 2016.

[55] A. P. Alves Da Silva and L. S. Moulin, "Confidence intervals for neural network based short-term load forecasting," *IEEE Trans. Power Syst.*, vol. 15, no. 4, pp. 1191–1196, 2000.

[56] M. Beccali, M. Cellura, V. Lo Brano, and A. Marvuglia, "Short-term prediction of household electricity consumption: Assessing weather sensitivity in a Mediterranean area," *Renew. Sustain. Energy Rev.*, vol. 12, no. 8, pp. 2040–2065, 2008.

[57] W. He, "Load Forecasting via Deep Neural Networks," *Procedia Comput. Sci.*, vol. 122, pp. 308–314, 2017.

[58] M. Hossain, S. Mekhilef, M. Danesh, L. Olatomiwa, and S. Shamshirband, "Application of extreme learning machine for short term output power forecasting of three grid-connected PV systems," *J. Clean. Prod.*, vol. 167, pp. 395–405, 2017.

[59] S. Sepasi, E. Reihani, A. M. Howlader, L. R. Roose, and M. M. Matsuura, "Very short term load forecasting of a distribution system with high PV penetration," *Renew. Energy*, vol. 106, pp. 142–148, 2017.

[60] F. Abbas, D. Feng, S. Habib, U. Rahman, A. Rasool, and Z. Yan, "Short term residential load forecasting: An improved optimal nonlinear auto regressive (NARX) method with exponential weight decay function," *Electron.*, vol. 7, no. 12, pp. 1–27, 2018.

[61] B. Yildiz, J. I. Bilbao, J. Dore, and A. B. Sproul, "Short-term forecasting of individual household electricity loads with investigating impact of data resolution and forecast horizon," *Renew. Energy Environ. Sustain.*, vol. 3, no. 3, p. 3, 2018.

[62] P. H. Kuo and C. J. Huang, "A high precision artificial neural networks model for short-Term energy load forecasting," *Energies*, vol. 11, no. 1, pp. 1–13, 2018.

[63] Z. Guo, K. Zhou, X. Zhang, and S. Yang, "A deep learning model for short-term power load and probability density forecasting," *Energy*, vol. 160, pp. 1186–1200, 2018.

[64] J. Bedi and D. Toshniwal, "Deep learning framework to forecast electricity demand," *Appl. Energy*, vol. 238, pp. 1312–1326, 2019.

[65] Y. Yang, W. Hong, and S. Li, "Deep ensemble learning based probabilistic load forecasting in smart grids," *Energy*, vol. 189, p. 116324, 2019.

[66] A. Safdarian, M. Fotuhi-Firuzabad, and M. Lehtonen, "A stochastic framework for short-term operation of a distribution company," *IEEE Trans. Power Syst.*, vol. 28, no. 4, pp. 4712–4721, 2013.

[67] V. S. Kodogiannis and E. M. Anagnostakis, "Soft computing based techniques for short-term load forecasting," *Fuzzy Sets Syst.*, vol. 128, no. 3, pp. 413–426, 2002.

[68] P. A. Mastorocostas, J. B. Theocharis, and V. S. Petridis, "A constrained orthogonal least-squares method for generating TSK fuzzy models: Application to short-term load forecasting," *Fuzzy Sets Syst.*, vol. 118, no. 2, pp. 215–233, 2001.

[69] J. Wang, R. Jia, W. Zhao, J. Wu, and Y. Dong, "Application of the largest Lyapunov exponent and non-linear fractal extrapolation algorithm to short-term load forecasting," *Chaos, Solitons and Fractals*, vol. 45, no. 9–10, pp. 1277–1287, 2012.

[70] K. Afshar and N. Bigdeli, "Data analysis and short term load forecasting in Iran electricity market using singular spectral analysis (SSA)," *Energy*, vol. 36, no. 5, pp. 2620–2627, 2011.

[71] A. E. Clements, A. S. Hurn, and Z. Li, "Forecasting day-ahead electricity load using a multiple equation time series approach," *Eur. J. Oper. Res.*, vol. 251, no. 2, pp. 522–530, 2016.

[72] L. J. Soares and L. R. Souza, "Forecasting electricity demand using generalized long memory," *Int. J. Forecast.*, vol. 22, no. 1, pp. 17–28, 2006.

[73] N. Liu, V. Babushkin, and A. Afshari, "Short-Term Forecasting of Temperature Driven Electricity Load Using Time Series and Neural Network Model," *J. Clean Energy Technol.*, vol. 2, no. 4, pp. 327–331, 2014.

[74] N. Amjady, "Short-term hourly load forecasting using time-series modeling with peak load estimation capability," *IEEE Trans. Power Syst.*, vol. 16, no. 4, pp. 798–805, 2001.

[75] E. Paparoditis and T. Sapatinas, "Short-Term Load Forecasting: The Similar Shape Functional Time Series Predictor," *IEEE Trans. Power Syst.*, vol. 28, no. 4, p. 22, 2012.

[76] M. Espinoza, C. Joye, R. Belmans, and B. De Moor, "Short-term load forecasting, profile identification, and customer segmentation: A methodology based on periodic time series," *IEEE Trans. Power Syst.*, vol. 20, no. 3, pp. 1622–1630, 2005.

[77] R. E. De Grande, A. Boukerche, and R. Alkharboush, "Time Series-Oriented Load Prediction Model and Migration Policies for Distributed Simulation Systems," *IEEE Trans. Parallel Distrib. Syst.*, vol. 28, no. 1, pp. 215–229, 2017.

[78] C. García-Ascanio and C. Maté, "Electric power demand forecasting using interval time series: A comparison between VAR and iMLP," *Energy Policy*, vol. 38, no. 2, pp. 715–725, 2010.

[79] D. H. Vu, K. M. Muttaqi, A. P. Agalgaonkar, and A. Bouzerdoum, "Short-term electricity demand forecasting using autoregressive based time varying model incorporating representative data adjustment," *Appl. Energy*, vol. 205, pp. 790–801, 2017.

[80] H. Sheng, J. Xiao, Y. Cheng, Q. Ni, and S. Wang, "Short-Term Solar Power Forecasting Based on Weighted Gaussian Process Regression," *IEEE Trans. Ind. Electron.*, vol. 65, no. 1, pp. 300–308, 2018.

[81] D. W. van der Meer, M. Shepero, A. Svensson, J. Widén, and J. Munkhammar, "Probabilistic forecasting of electricity consumption, photovoltaic power generation and net demand of an individual building using Gaussian Processes," *Appl. Energy*, vol. 213, pp. 195–207, 2018.

[82] J. Cordova, L. M. K. Sriram, A. Kocatepe, Y. Zhou, E. E. Ozguven, and R. Arghandeh, "Combined Electricity and Traffic Short-Term Load Forecasting Using Bundled Causality Engine," *IEEE Trans. Intell. Transp. Syst.*, vol. 20, no. 9, pp. 3448–3458, 2019.

[83] A. M. Al-Kandari, S. A. Soliman, and M. E. El-Hawary, "Fuzzy short-term electric load forecasting," *Int. J. Electr. Power Energy Syst.*, vol. 26, no. 2, pp. 111–122, 2004.

[84] S. Chenthur Pandian, K. Duraiswamy, C. C. A. Rajan, and N. Kanagaraj, "Fuzzy approach for short term load forecasting," *Electr. Power Syst. Res.*, vol. 76, no. 6–7, pp. 541–548, 2006.

[85] M. Tamimi and R. Egbert, "Short term electric load forecasting via fuzzy neural collaboration," *Electr. Power Syst. Res.*, vol. 56, no. 3, pp. 243–248, 2000.

[86] R. H. Liang and C. C. Cheng, "Short-term load forecasting by a neuro-fuzzy based approach," *Int. J. Electr. Power Energy Syst.*, vol. 24, no. 2, pp. 103–111, 2002.

[87] G. Zahedi, S. Azizi, A. Bahadori, A. Elkamel, and S. R. Wan Alwi, "Electricity demand estimation using an adaptive neuro-fuzzy network: A case study from the Ontario province - Canada," *Energy*, vol. 49, no. 1, pp. 323–328, 2013.

[88] R. Mamlook, O. Badran, and E. Abdulhadi, "A fuzzy inference model for short-term load forecasting," *Energy Policy*, vol. 37, no. 4, pp. 1239–1248, 2009.

[89] S. E. Papadakis, J. B. Theocharis, and A. G. Bakirtzis, "A load curve based fuzzy modeling technique for short-term load forecasting," *Fuzzy Sets Syst.*, vol. 135, no. 2, pp. 279–303, 2003.

[90] S. J. Huang and K. R. Shih, "Application of a fuzzy model for short-term load forecast with group method of data handling enhancement," *Int. J. Electr. Power Energy Syst.*, vol. 24, no. 8, pp. 631–638, 2002.

[91] H. Y. Yang, H. Ye, G. Wang, J. Khan, and T. Hu, "Fuzzy neural very-short-term load forecasting based on chaotic dynamics reconstruction," *Chaos, Solitons and Fractals*, vol. 29, no. 2, pp. 462–469, 2006.

[92] S. Kucukali and K. Baris, "Turkey's short-term gross annual electricity demand forecast by fuzzy logic approach," *Energy Policy*, vol. 38, no. 5, pp. 2438–2445, 2010.

[93] R. Efendi, Z. Ismail, and M. M. Deris, "A new linguistic out-sample approach of fuzzy time series for daily forecasting of Malaysian electricity load demand," *Appl. Soft Comput. J.*, vol. 28, pp. 422–430, 2015.

[94] M. Karimi, H. Karami, M. Gholami, H. Khatibzadehazad, and N. Moslemi, "Priority index considering temperature and date proximity for selection of similar days in knowledge-based short term load forecasting method," *Energy*, vol. 144, pp. 928–940, 2018.

[95] S. S. Pappas *et al.*, "Electricity demand load forecasting of the Hellenic power system using an ARMA model," *Electr. Power Syst. Res.*, vol. 80, no. 3, pp. 256–264, 2010.

[96] J. W. Taylor, "An evaluation of methods for very short-term load forecasting using minute-by-minute British data," *Int. J. Forecast.*, vol. 24, no. 4, pp. 645–658, 2008.

[97] A. K. Topalli, I. Erkmen, and I. Topalli, "Intelligent short-term load forecasting in Turkey," *Int. J. Electr. Power Energy Syst.*, vol. 28, no. 7, pp. 437–447, 2006.

[98] Y. Wang, J. Wang, G. Zhao, and Y. Dong, "Application of residual modification approach in seasonal ARIMA for electricity demand forecasting: A case study of China," *Energy Policy*, vol. 48, pp. 284–294, 2012.

[99] C. M. Lee and C. N. Ko, "Short-term load forecasting using lifting scheme and ARIMA models," *Expert Syst. Appl.*, vol. 38, no. 5, pp. 5902–5911, 2011.

[100] M. Matijaš, J. A. K. Suykens, and S. Krajcar, "Load forecasting using a multivariate meta-learning system," *Expert Syst. Appl.*, vol. 40, no. 11, pp. 4427–4437, 2013.

[101] J. W. Taylor, "Short-term electricity demand forecasting using double seasonal exponential smoothing," *J. Oper. Res. Soc.*, vol. 54, no. 8, pp. 799–805, 2003.

[102] A. Tarsitano and I. L. Amerise, "Short-term load forecasting using a two-stage sarimax model," *Energy*, vol. 133, pp. 108–114, 2017.

[103] N. Elamin and M. Fukushige, "Modeling and forecasting hourly electricity demand by SARIMAX with interactions," *Energy*, vol. 165, pp. 257–268, 2018.

[104] C. Zhou and X. Chen, "Predicting energy consumption: A multiple decomposition-ensemble approach," *Energy*, vol. 189, p. 116045, 2019.

[105] E. Almeshaiei and H. Soltan, "A methodology for Electric Power Load Forecasting," *Alexandria Eng. J.*, vol. 50, no. 2, pp. 137–144, 2011.

[106] J. W. Taylor, "Short-term load forecasting with exponentially weighted methods," *IEEE Trans. Power Syst.*, vol. 27, no. 1, pp. 458–464, 2012.

[107] B. A. Hoverstad, A. Tidemann, H. Langseth, and P. Ozturk, "Short-Term Load Forecasting With Seasonal Decomposition Using Evolution for Parameter Tuning," *IEEE Trans. Smart Grid*, vol. 6, no. 4, pp. 1904–1913, 2015.

[108] X. Zhang and J. Wang, "A novel decomposition-ensemble model for forecasting short-term load-time series with multiple seasonal patterns," *Appl. Soft Comput. J.*, vol. 65, pp. 478–494, 2018.

[109] G. D. Li, C. H. Wang, S. Masuda, and M. Nagai, "A research on short term load forecasting problem applying improved grey dynamic model," *Int. J. Electr. Power Energy Syst.*, vol. 33, no. 4, pp. 809–816, 2011.

[110] M. Jin, X. Zhou, Z. M. Zhang, and M. M. Tentzeris, "Short-term power load forecasting using grey correlation contest modeling," *Expert Syst. Appl.*, vol. 39, no. 1, pp. 773–779, 2012.

[111] K. Li and T. Zhang, "Forecasting electricity consumption using an improved grey prediction model," *Inf.*, vol. 9, no. 8, pp. 1–18, 2018.

[112] Q. Wang, S. Li, and R. Li, "Forecasting energy demand in China and India: Using single-linear, hybrid-linear, and non-linear time series forecast techniques," *Energy*, vol. 161, pp. 821–831, 2018.

[113] C. Guan, P. B. Luh, L. D. Michel, and Z. Chi, "Hybrid Kalman filters for very short-term load forecasting and prediction interval estimation," *IEEE Trans. Power Syst.*, vol. 28, no. 4, pp. 3806–3817, 2013.

[114] H. M. Al-Hamadi and S. A. Soliman, "Short-term electric load forecasting based on Kalman filtering algorithm with moving window weather and load model," *Electr. Power Syst. Res.*, vol. 68, no. 1, pp. 47–59, 2004.

[115] C. M. Huang, C. J. Huang, and M. L. Wang, "A particle swarm optimization to identifying the ARMAX model for short-term load forecasting," *IEEE Trans. Power Syst.*, vol. 20, no. 2, pp. 1126–1133, 2005.

[116] M. El-Telbany and F. El-Karmi, "Short-term forecasting of Jordanian electricity demand using particle swarm optimization," *Electr. Power Syst. Res.*, vol. 78, no. 3, pp. 425–433, 2008.

[117] O. A. S. Carpinteiro and A. P. Alves Da Silva, "A hierarchical self-organizing map model in short-term load forecasting," *J. Intell. Robot. Syst. Theory Appl.*, vol. 31, no. 1–3, pp. 105–113, 2001.

[118] M. López, S. Valero, C. Senabre, J. Aparicio, and A. Gabaldon, "Application of SOM neural networks to short-term load forecasting: The Spanish electricity market case study," *Electr. Power Syst. Res.*, vol. 91, pp. 18–27, 2012.

[119] H. Mao, X. J. Zeng, G. Leng, Y. J. Zhai, and J. A. Keane, "Short-term and midterm load forecasting using a bilevel optimization model," *IEEE Trans. Power Syst.*, vol. 24, no. 2, pp. 1080–1090, 2009.

[120] M. D. Toksari, "Estimating the net electricity energy generation and demand using the ant colony optimization approach: Case of Turkey," *Energy Policy*, vol. 37, no. 3, pp. 1181–1187, 2009.

[121] S. H. Ling, F. H. F. Leung, H. K. Lam, Y. S. Lee, and P. K. S. Tam, "A novel genetic-algorithm-based neural network for short-term load forecasting," *IEEE Trans. Ind. Electron.*, vol. 50, no. 4, pp. 793–799, 2003.

[122] K. Nose-Filho, A. D. P. Lotufo, and C. R. Minussi, "Short-term multinodal load forecasting using a modified general regression neural network," *IEEE Trans. Power Deliv.*, vol. 26, no. 4, pp. 2862–2869, 2011.

[123] M. Alamaniotis, A. Ikonomopoulos, and L. H. Tsoukalas, "Evolutionary multiobjective optimization of kernel-based very-short-term load forecasting," *IEEE Trans. Power Syst.*, vol. 27, no. 3, pp. 1477–1484, 2012.

[124] J. M. Vilar, R. Cao, and G. Aneiros, "Forecasting next-day electricity demand and price using nonparametric functional methods," *Int. J. Electr. Power Energy Syst.*, vol. 39, no. 1, pp. 48–55, 2012.

[125] J. R. Cancelo, A. Espasa, and R. Grafe, "Forecasting the electricity load from one day to one week ahead for the Spanish system operator," *Int. J. Forecast.*, vol. 24, no. 4, pp. 588–602, 2008.

[126] N. Charlton and C. Singleton, "A refined parametric model for short term load forecasting," *Int. J. Forecast.*, vol. 30, no. 2, pp. 364–368, 2014.

[127] M. Zamo, O. Mestre, P. Arbogast, and O. Pannekoucke, "A benchmark of statistical regression methods for short-term forecasting of photovoltaic electricity production, part I: Deterministic forecast of hourly production," *Sol. Energy*, vol. 105, pp. 792–803, 2014.

[128] C. A. Maia and M. M. Gonçalves, "A methodology for short-term electric load forecasting based on specialized recursive digital filters," *Comput. Ind. Eng.*, vol. 57, no. 3, pp. 724–731, 2009.

[129] L. F. Amaral, R. C. Souza, and M. Stevenson, "A smooth transition periodic autoregressive (STPAR) model for short-term load forecasting," *Int. J. Forecast.*, vol. 24, no. 4, pp. 603–615, 2008.

[130] L. J. Soares and M. C. Medeiros, "Modeling and forecasting short-term electricity load: A comparison of methods with an application to Brazilian data," *Int. J. Forecast.*, vol. 24, no. 4, pp. 630–644, 2008.

[131] S. Ružić, A. Vučković, and N. Nikolić, "Weather Sensitive Method for Short Term Load Forecasting in Electric Power Utility of Serbia," *IEEE Trans. Power Syst.*, vol. 18, no. 4, pp. 1581–1586, 2003.

[132] K. Bin Song, Y. S. Baek, D. H. Hong, and G. Jang, "Short-term load forecasting for the holidays using fuzzy linear regression method," *IEEE Trans. Power Syst.*, vol. 20, no. 1, pp. 96–101, 2005.

[133] G. Dudek, "Pattern-based local linear regression models for short-term load forecasting," *Electr. Power Syst. Res.*, vol. 130, pp. 139–147, 2016.

[134] J. Yang and J. Stenzel, "Short-term load forecasting with increment regression tree," *Electr. Power Syst. Res.*, vol. 76, no. 9–10, pp. 880–888, 2006.

[135] S. Fan and R. J. Hyndman, "Short-term load forecasting based on a semi-parametric additive model," *IEEE Trans. Power Syst.*, vol. 27, no. 1, pp. 134–141, 2012.

[136] N. J. Johannesen, M. Kolhe, and M. Goodwin, "Relative evaluation of regression tools for urban area electrical energy demand forecasting," *J. Clean. Prod.*, vol. 218, pp. 555–564, 2019.

[137] Y. Yang, S. Li, W. Li, and M. Qu, "Power load probability density forecasting using Gaussian process quantile regression," *Appl. Energy*, vol. 213, pp. 499–509, 2018.

[138] X. Fu, X. J. Zeng, P. Feng, and X. Cai, "Clustering-based short-term load forecasting for residential electricity under the increasing-block pricing tariffs in China," *Energy*, vol. 165, pp. 76–89, 2018.

[139] K. Nagbe, J. Cugliari, and J. Jacques, "Short-term electricity demand forecasting using a functional state space model," *Energies*, vol. 11, no. 5, pp. 1–24, 2018.

[140] M. E. Lebotsa, C. Sigauke, A. Bere, R. Fildes, and J. E. Boylan, "Short term electricity demand forecasting using partially linear additive quantile regression with an application to the unit commitment problem," *Appl. Energy*, vol. 222, pp. 104–118, 2018.

[141] B. Uniejewski, G. Marcjasz, and R. Weron, "Understanding intraday electricity markets: Variable selection and very short-term price forecasting using LASSO," *Int. J. Forecast.*, vol. 35, no. 4, pp. 1533–1547, 2019.

[142] Ó. Trull, J. C. García-Díaz, and A. Troncoso, "Application of discrete-interval moving seasonalities to Spanish electricity demand forecasting during easter," *Energies*, vol. 12, no. 6, pp. 1–16, 2019.

[143] Y. Wang and J. M. Bielicki, "Acclimation and the response of hourly electricity loads to meteorological variables," *Energy*, vol. 142, pp. 473–485, 2018.

[144] P. Alipour, S. Mukherjee, and R. Nateghi, "Assessing climate sensitivity of peak electricity load for resilient power systems planning and operation: A study applied to the Texas region," *Energy*, vol. 185, pp. 1143–1153, 2019.

[145] K. Kavaklioglu, "Modeling and prediction of Turkey's electricity consumption using Support Vector Regression," *Appl. Energy*, vol. 88, no. 1, pp. 368–375, 2011.

[146] W. C. Hong, "Electric load forecasting by support vector model," *Appl. Math. Model.*, vol. 33, no. 5, pp. 2444–2454, 2009.

[147] G. F. Fan, L. L. Peng, W. C. Hong, and F. Sun, "Electric load forecasting by the SVR model with differential empirical mode decomposition and auto regression," *Neurocomputing*, vol. 173, pp. 958–970, 2016.

[148] J. Wang, W. Zhu, W. Zhang, and D. Sun, "A trend fixed on firstly and seasonal adjustment model combined with the ε-SVR for short-term forecasting of electricity demand," *Energy Policy*, vol. 37, no. 11, pp. 4901–4909, 2009.

[149] D. Niu, Y. Wang, and D. D. Wu, "Power load forecasting using support vector machine and ant colony optimization," *Expert Syst. Appl.*, vol. 37, no. 3, pp. 2531–2539, 2010.

[150] Y. Ren, P. N. Suganthan, N. Srikanth, and G. Amaratunga, "Random vector functional link network for short-term electricity load demand forecasting," *Inf. Sci. (Ny).*, vol. 367–368, pp. 1078–1093, 2016.

[151] Z. Hu, Y. Bao, T. Xiong, and R. Chiong, "Hybrid filter-wrapper feature selection for short-term load forecasting," *Eng. Appl. Artif. Intell.*, vol. 40, pp. 17–27, 2015.

[152] Y. Li, J. Che, and Y. Yang, "Subsampled support vector regression ensemble for short term electric load forecasting," *Energy*, vol. 164, pp. 160–170, 2018.

[153] A. Afshari and L. A. Friedrich, "Inverse modeling of the urban energy system using hourly electricity demand and weather measurements, Part 1: Black-box model," *Energy Build.*, vol. 157, pp. 126–138, 2017.

[154] B. P. Hayes, J. K. Gruber, and M. Prodanovic, "Multi-nodal short-term energy forecasting using smart meter data," *IET Gener. Transm. Distrib.*, vol. 12, no. 12, pp. 2988–2994, 2018.

[155] K. Chapagain and S. Kittipiyakul, "Performance analysis of short-term electricity demand with atmospheric variables," *Energies*, vol. 11, no. 4, pp. 1–34, 2018.

[156] D. X. Niu, H. F. Shi, and D. D. Wu, "Short-term load forecasting using bayesian neural networks learned by Hybrid Monte Carlo algorithm," *Appl. Soft Comput. J.*, vol. 12, no. 6, pp. 1822–1827, 2012.

[157] D. K. Chaturvedi, M. Mohan, R. K. Singh, and P. K. Kalra, "Improved generalized neuron model for short-term load forecasting," *Soft Comput.*, vol. 8, no. 1, pp. 10–18, 2003.

[158] G. Cervone, L. Clemente-Harding, S. Alessandrini, and L. Delle Monache, "Short-term photovoltaic power forecasting using Artificial Neural Networks and an Analog Ensemble," *Renew. Energy*, vol. 108, pp. 274–286, 2017.

[159] A. Kheirkhah, A. Azadeh, M. Saberi, A. Azaron, and H. Shakouri, "Improved estimation of electricity demand function by using of artificial neural network, principal component analysis and data envelopment analysis," *Comput. Ind. Eng.*, vol. 64, no. 1, pp. 425–441, 2013.

[160] Y. T. Chae, R. Horesh, Y. Hwang, and Y. M. Lee, "Artificial neural network model for forecasting sub-hourly electricity usage in commercial buildings," *Energy Build.*, vol. 111, pp. 184–194, 2016.

[161] A. Anand and L. Suganthi, "Hybrid GA-PSO optimization of Artificial Neural Network for forecasting electricity demand," *Energies*, vol. 11, no. 4, pp. 1–15, 2018.

[162] A. Ahmad, N. Javaid, A. Mateen, M. Awais, and Z. A. Khan, "Short-Term load forecasting in smart grids: An intelligent modular approach," *Energies*, vol. 12, no. 1, pp. 1–21, 2019.

[163] J. L. Casteleiro-Roca *et al.*, "Short-term energy demand forecast in hotels using hybrid intelligent modeling," *Sensors (Switzerland)*, vol. 19, no. 11, pp. 1–18, 2019.

[164] M. Ghofrani, M. Ghayekhloo, A. Arabali, and A. Ghayekhloo, "A hybrid short-term load forecasting with a new input selection framework," *Energy*, vol. 81, pp. 777–786, 2015.

[165] J. Kim, J. Moon, E. Hwang, and P. Kang, "Recurrent inception convolution neural network for multi short-term load forecasting," *Energy Build.*, vol. 194, pp. 328–341, 2019.

[166] A. S. Pandey, D. Singh, and S. K. Sinha, "Intelligent hybrid wavelet models for short-term load forecasting," *IEEE Trans. Power Syst.*, vol. 25, no. 3, pp. 1266–1273, 2010.

[167] A. T. Eseye, M. Lehtonen, T. Tukia, S. Uimonen, and R. John Millar, "Machine learning based integrated feature selection approach for improved electricity demand forecasting in decentralized energy systems," *IEEE Access*, vol. 7, pp. 91463–91475, 2019.

[168] K. Yan, X. Wang, Y. Du, N. Jin, H. Huang, and H. Zhou, "Multi-step short-term power consumption forecasting with a hybrid deep learning strategy," *Energies*, vol. 11, no. 11, pp. 1–15, 2018.

[169] H. J. Sadaei, P. C. de Lima e Silva, F. G. Guimarães, and M. H. Lee, "Short-term load forecasting by using a combined method of convolutional neural networks and fuzzy time series," *Energy*, vol. 175, pp. 365–377, 2019.

[170] S. M. Kelo and S. V. Dudul, "Short-term Maharashtra state electrical power load prediction with special emphasis on seasonal changes using a novel focused time lagged recurrent neural network based on time delay neural network model," *Expert Syst. Appl.*, vol. 38, no. 3, pp. 1554–1564, 2011.

[171] G. C. Liao and T. P. Tsao, "Application of a fuzzy neural network combined with a chaos genetic algorithm and simulated annealing to short-term load forecasting," *IEEE Trans. Evol. Comput.*, vol. 10, no. 3, pp. 330–340, 2006.

[172] D. C. Li, C. J. Chang, C. C. Chen, and W. C. Chen, "Forecasting short-term electricity consumption using the adaptive grey-based approach-An Asian case," *Omega*, vol. 40, no. 6, pp. 767–773, 2012.

[173] T. Senjyu, P. Mandal, K. Uezato, and T. Funabashi, "Next day load curve forecasting using hybrid correction method," *IEEE Trans. Power Syst.*, vol. 20, no. 1, pp. 102–109, 2005.

[174] M. Kim, W. Choi, Y. Jeon, and L. Liu, "A hybrid neural network model for power demand forecasting," *Energies*, vol. 12, no. 5, pp. 1–17, 2019.

[175] K. H. Kim, H. S. Youn, and Y. C. Kang, "Short-term load forecasting for special days in anomalous load conditions using neural networks and fuzzy inference method," *IEEE Trans. Power Syst.*, vol. 15, no. 2, pp. 559–565, 2000.

[176] L. Liu *et al.*, "Prediction of short-term PV power output and uncertainty analysis," *Appl. Energy*, vol. 228, pp. 700–711, 2018.

[177] Z. Yun, Z. Quan, S. Caixin, L. Shaolan, L. Yuming, and S. Yang, "RBF neural network and ANFIS-based short-term load forecasting approach in real-time price environment," *IEEE Trans. Power Syst.*, vol. 23, no. 3, pp. 853–858, 2008.

[178] E. Sala-Cardoso, M. Delgado-Prieto, K. Kampouropoulos, and L. Romeral, "Activity-aware HVAC power demand forecasting," *Energy Build.*, vol. 170, pp. 15–24, 2018.

[179] H. Chitsaz, H. Shaker, H. Zareipour, D. Wood, and N. Amjady, "Short-term electricity load forecasting of buildings in microgrids," *Energy Build.*, vol. 99, pp. 50–60, Apr. 2015.

[180] S. Kulkarni, S. P. Simon, and K. Sundareswaran, "A spiking neural network (SNN) forecast engine for short-term electrical load forecasting," *Appl. Soft Comput. J.*, vol. 13, no. 8, pp. 3628–3635, 2013.

[181] X. Qiu, P. N. Suganthan, and G. A. J. Amaratunga, "Ensemble incremental learning Random Vector Functional Link network for short-term electric load forecasting," *Knowledge-Based Syst.*, vol. 145, pp. 1–14, 2018.

[182] D. K. Chaturvedi, A. P. Sinha, and O. P. Malik, "Short term load forecast using fuzzy logic and wavelet transform integrated generalized neural network," *Int. J. Electr. Power Energy Syst.*, vol. 67, pp. 230–237, 2015.

[183] B. L. Zhang and Z. Y. Dong, "An adaptive neural-wavelet model for short term load forecasting," *Electr. Power Syst. Res.*, vol. 59, no. 2, pp. 121–129, 2001.

[184] A. Deihimi, O. Orang, and H. Showkati, "Short-term electric load and temperature forecasting using wavelet echo state networks with neural reconstruction," *Energy*, vol. 57, pp. 382–401, 2013.

[185] R. A. Hooshmand, H. Amooshahi, and M. Parastegari, "A hybrid intelligent algorithm based short-term load forecasting approach," *Int. J. Electr. Power Energy Syst.*, vol. 45, no. 1, pp. 313–324, 2013.

[186] S. Li, P. Wang, and L. Goel, "Short-term load forecasting by wavelet transform and evolutionary extreme learning machine," *Electr. Power Syst. Res.*, vol. 122, pp. 96–103, 2015.

[187] S. Bahrami, R. A. Hooshmand, and M. Parastegari, "Short term electric load forecasting by wavelet transform and grey model improved by PSO (particle swarm optimization) algorithm," *Energy*, vol. 72, no. 2014, pp. 434–442, 2014.

[188] N. Amjady and F. Keynia, "Short-term load forecasting of power systems by combination of wavelet transform and neuro-evolutionary algorithm," *Energy*, vol. 34, no. 1, pp. 46–57, 2009.

[189] S. Li, L. Goel, and P. Wang, "An ensemble approach for short-term load forecasting by extreme learning machine," *Appl. Energy*, vol. 170, pp. 22–29, 2016.

[190] G. Sudheer and A. Suseelatha, "Short term load forecasting using wavelet transform combined with Holt-Winters and weighted nearest neighbor models," *Int. J. Electr. Power Energy Syst.*, vol. 64, pp. 340–346, 2015.

[191] H. T. Nguyen and I. T. Nabney, "Short-term electricity demand and gas price forecasts using wavelet transforms and adaptive models," *Energy*, vol. 35, no. 9, pp. 3674–3685, 2010.

[192] M. Salami, F. M. Sobhani, and M. S. Ghazizadeh, "Short-term forecasting of electricity supply and demand by using the wavelet-PSO-NNS-SO technique for searching in big data of iran's electricity market," *Data*, vol. 3, no. 4, pp. 1–26, 2018.

[193] C. Il Kim, I. K. Yu, and Y. H. Song, "Kohonen neural network and wavelet transform based approach to short-term load forecasting," *Electr. Power Syst. Res.*, vol. 63, no. 3, pp. 169–176, 2002.

[194] N. V. Truong, L. Wang, and P. K. C. Wong, "Modelling and short-term forecasting of daily peak power demand in Victoria using two-dimensional wavelet based SDP models," *Int. J. Electr. Power Energy Syst.*, vol. 30, no. 9, pp. 511–518, 2008.

[195] A. J. da Rocha Reis and A. P. Alves da Silva, "Feature extraction via multiresolution analysis for short-term load forecasting," *IEEE Trans. Power Syst.*, vol. 20, no. 1, pp. 189–198, 2005.

[196] A. Selakov, D. Cvijetinović, L. Milović, S. Mellon, and D. Bekut, "Hybrid PSO-SVM method for short-term load forecasting during periods with significant temperature variations in city of Burbank," *Appl. Soft Comput. J.*, vol. 16, pp. 80–88, 2014.

[197] B. Wang, N. ling Tai, H. qing Zhai, J. Ye, J. dong Zhu, and L. bo Qi, "A new ARMAX model based on evolutionary algorithm and particle swarm optimization for short-term load forecasting," *Electr. Power Syst. Res.*, vol. 78, no. 10, pp. 1679–1685, 2008.

[198] Z. Hu, Y. Bao, and T. Xiong, "Comprehensive learning particle swarm optimization based memetic algorithm for model selection in short-term load forecasting using support vector regression," *Appl. Soft Comput. J.*, vol. 25, pp. 15–25, 2014.

[199] P. Duan, K. Xie, T. Guo, and X. Huang, "Short-Term Load Forecasting for Electric Power Systems Using the PSO-SVR and FCM Clustering Techniques," *Energies*, vol. 4, no. 1, pp. 173–184, 2011.

[200] M. Dadkhah, M. Jahangoshai Rezaee, and A. Zare Chavoshi, "Short-term power output forecasting of hourly operation in power plant based on climate factors and effects of wind direction and wind speed," *Energy*, vol. 148, pp. 775–788, 2018.

[201] S. Fan and L. Chen, "Short-term load forecasting based on an adaptive hybrid method," *IEEE Trans. Power Syst.*, vol. 21, no. 1, pp. 392–401, 2006.

[202] L. Hernández, C. Baladrón, J. M. Aguiar, B. Carro, and A. Sánchez-Esguevillas, "Classification and clustering of electricity demand patterns in industrial parks," *Energies*, vol. 5, no. 12, pp. 5215–5228, 2012.

[203] M. R. Amin-Naseri and A. R. Soroush, "Combined use of unsupervised and supervised learning for daily peak load forecasting," *Energy Convers. Manag.*, vol. 49, no. 6, pp. 1302–1308, 2008.

[204] H. Shakouri G., R. Nadimi, and F. Ghaderi, "A hybrid TSK-FR model to study short-term variations of the electricity demand versus the temperature changes," *Expert Syst. Appl.*, vol. 36, no. 2 PART 1, pp. 1765–1772, 2009.

[205] A. Ghanbari, S. M. R. Kazemi, F. Mehmanpazir, and M. M. Nakhostin, "A cooperative ant colony optimization-genetic algorithm approach for construction of energy demand forecasting knowledge-based expert systems," *Knowledge-Based Syst.*, vol. 39, pp. 194–206, 2013.

[206] M. Niu, S. Sun, J. Wu, L. Yu, and J. Wang, "An innovative integrated model using the singular spectrum analysis and nonlinear multi-layer perceptron network optimized by hybrid intelligent algorithm for short-term load forecasting," *Appl. Math. Model.*, vol. 40, no. 5–6, pp. 4079–4093, 2016.

[207] F. Amara, K. Agbossou, Y. Dubé, S. Kelouwani, A. Cardenas, and S. S. Hosseini, "A residual load modeling approach for household short-term load forecasting application," *Energy Build.*, vol. 187, pp. 132–143, 2019.

[208] K. G. Boroojeni, M. H. Amini, S. Bahrami, S. S. Iyengar, A. I. Sarwat, and O. Karabasoglu, "A novel multi-time-scale modeling for electric power demand forecasting: From short-term to medium-term horizon," *Electr. Power Syst. Res.*, vol. 142, pp. 58–73, 2017.

[209] Q. Zhou, L. Tesfatsion, and C. C. Liu, "Short-term congestion forecasting in wholesale power markets," *IEEE Trans. Power Syst.*, vol. 26, no. 4, pp. 2185–2196, 2011.

[210] S. Zhu, J. Wang, W. Zhao, and J. Wang, "A seasonal hybrid procedure for electricity demand forecasting in China," *Appl. Energy*, vol. 88, no. 11, pp. 3807–3815, 2011.

[211] J. Wang, D. Chi, J. Wu, and H. Y. Lu, "Chaotic time series method combined with particle swarm optimization and trend adjustment for electricity demand forecasting," *Expert Syst. Appl.*, vol. 38, no. 7, pp. 8419–8429, 2011.

[212] Y. Wang, Q. Xia, and C. Kang, "Secondary forecasting based on deviation analysis for short-term load forecasting," *IEEE Trans. Power Syst.*, vol. 26, no. 2, pp. 500–507, 2011.

[213] F. M. Bianchi, E. De Santis, A. Rizzi, and A. Sadeghian, "Short-Term Electric Load Forecasting Using Echo State Networks and PCA Decomposition," *IEEE Access*, vol. 3, pp. 1931–1943, 2015.

[214] N. Liu, Q. Tang, J. Zhang, W. Fan, and J. Liu, "A hybrid forecasting model with parameter optimization for short-term load forecasting of micro-grids," *Appl. Energy*, vol. 129, pp. 336–345, 2014.

[215] M. Barman and N. B. Dev Choudhury, "Season specific approach for short-term load forecasting based on hybrid FA-SVM and similarity concept," *Energy*, vol. 174, pp. 886–896, 2019.

[216] N. Amjady, "Short-term bus load forecasting of power systems by a new hybrid method," *IEEE Trans. Power Syst.*, vol. 22, no. 1, pp. 333–341, 2007.

[217] V. H. Hinojosa and A. Hoese, "Short-term load forecasting using fuzzy inductive reasoning and evolutionary algorithms," *IEEE Trans. Power Syst.*, vol. 25, no. 1, pp. 565–574, 2010.

[218] D. Akay and M. Atak, "Grey prediction with rolling mechanism for electricity demand forecasting of Turkey," *Energy*, vol. 32, no. 9, pp. 1670–1675, 2007.

[219] R. Enayatifar, H. J. Sadaei, A. H. Abdullah, and A. Gani, "Imperialist competitive algorithm combined with refined high-order weighted fuzzy time series (RHWFTS-ICA) for short term load forecasting," *Energy Convers. Manag.*, vol. 76, pp. 1104–1116, 2013.

[220] F. Wang *et al.*, "Daily pattern prediction based classification modeling approach for day-ahead electricity price forecasting," *Int. J. Electr. Power Energy Syst.*, vol. 105, pp. 529–540, 2019.

[221] M. Khan *et al.*, "Game theoretical demand response management and short-term load forecasting by knowledge based systems on the basis of priority index," *Electron.*, vol. 7, no. 12, pp. 1–34, 2018.

[222] Y. Wang, Y. Shen, S. Mao, X. Chen, and H. Zou, "LASSO and LSTM integrated temporal model for short-term solar intensity forecasting," *IEEE Internet Things J.*, vol. 6, no. 2, pp. 2933–2944, 2019.

[223] A. Yang, W. Li, and X. Yang, "Short-term electricity load forecasting based on feature selection and Least Squares Support Vector Machines," *Knowledge-Based Syst.*, vol. 163, pp. 159–173, 2019.

[224] L. Bartolucci, S. Cordiner, V. Mulone, and M. Santarelli, "Short-therm forecasting method to improve the performance of a model predictive control strategy for a residential hybrid renewable energy system," *Energy*, vol. 172, pp. 997–1004, 2019.

[225] L. Xiao, W. Shao, T. Liang, and C. Wang, "A combined model based on multiple seasonal patterns and modified firefly

algorithm for electrical load forecasting," *Appl. Energy*, vol. 167, pp. 135–153, 2016.

[226] S. Kouhi, F. Keynia, and S. Najafi Ravadanegh, "A new short-term load forecast method based on neuro-evolutionary algorithm and chaotic feature selection," *Int. J. Electr. Power Energy Syst.*, vol. 62, pp. 862–867, 2014.

[227] H. Kebriaei, B. N. Araabi, and A. Rahimi-Kian, "Short-term load forecasting with a new nonsymmetric penalty function," *IEEE Trans. Power Syst.*, vol. 26, no. 4, pp. 1817–1825, 2011.

[228] H. J. Sadaei, R. Enayatifar, A. H. Abdullah, and A. Gani, "Short-term load forecasting using a hybrid model with a refined exponentially weighted fuzzy time series and an improved harmony search," *Int. J. Electr. Power Energy Syst.*, vol. 62, no. from 2005, pp. 118–129, 2014.

[229] G. Aneiros, J. Vilar, and P. Raña, "Short-term forecast of daily curves of electricity demand and price," *Int. J. Electr. Power Energy Syst.*, vol. 80, pp. 96–108, 2016.

[230] M. Barman, N. B. Dev Choudhury, and S. Sutradhar, "A regional hybrid GOA-SVM model based on similar day approach for short-term load forecasting in Assam, India," *Energy*, vol. 145, pp. 710–720, 2018.

[231] M. M. Tripathi, K. G. Upadhyay, and S. N. Singh, "Short-Term Load Forecasting Using Generalized Regression and Probabilistic Neural Networks in the Electricity Market," *Electr. J.*, vol. 21, no. 9, pp. 24–34, 2008.

[232] R. Hu, S. Wen, Z. Zeng, and T. Huang, "A short-term power load forecasting model based on the generalized regression neural network with decreasing step fruit fly optimization algorithm," *Neurocomputing*, vol. 221, no. September 2016, pp. 24–31, 2017.

[233] G. A. Darbellay and M. Slama, "Forecasting the short-term demand for electricity: Do neural networks stand a better chance?," *Int. J. Forecast.*, vol. 16, no. 1, pp. 71–83, 2000.

[234] Y. He, Q. Xu, J. Wan, and S. Yang, "Short-term power load probability density forecasting based on quantile regression neural network and triangle kernel function," *Energy*, vol. 114, pp. 498–512, 2016.

[235] A. Laouafi, M. Mordjaoui, S. Haddad, T. E. Boukelia, and A. Ganouche, "Online electricity demand forecasting based on an effective forecast combination methodology," *Electr. Power Syst. Res.*, vol. 148, pp. 35–47, 2017.

[236] V. Yadav and D. Srinivasan, "A SOM-based hybrid linear-neural model for short-term load forecasting," *Neurocomputing*, vol. 74, no. 17, pp. 2874–2885, 2011.

[237] F. Divina, A. Gilson, F. Goméz-Vela, M. G. Torres, and J. F. Torres, "Stacking ensemble learning for short-term electricity consumption forecasting," *Energies*, vol. 11, no. 4, pp. 1–31, 2018.

[238] D. Koschwitz, J. Frisch, and C. van Treeck, "Data-driven heating and cooling load predictions for non-residential buildings based on support vector machine regression and NARX Recurrent Neural Network: A comparative study on district scale," *Energy*, vol. 165, pp. 134–142, 2018.

[239] Y. Guo *et al.*, "Machine learning-based thermal response time ahead energy demand prediction for building heating systems," *Appl. Energy*, vol. 221, pp. 16–27, 2018.

[240] R. H. Liang and C. C. Cheng, "Combined regression-fuzzy approach for short-term load forecasting," *IEE Proc. Gener. Transm. Distrib.*, vol. 147, no. 4, pp. 261–266, 2000.

[241] J. Che, "A novel hybrid model for bi-objective short-term electric load forecasting," *Int. J. Electr. Power Energy Syst.*, vol. 61, pp. 259–266, 2014.

[242] G. F. Fan, L. L. Peng, and W. C. Hong, "Short term load forecasting based on phase space reconstruction algorithm and bi-square kernel regression model," *Appl. Energy*, vol. 224, pp. 13–33, 2018.

[243] J. Wu, J. Wang, H. Lu, Y. Dong, and X. Lu, "Short term load forecasting technique based on the seasonal exponential adjustment method and the regression model," *Energy Convers. Manag.*, vol. 70, pp. 1–9, 2013.

[244] C. N. Ko and C. M. Lee, "Short-term load forecasting using SVR (support vector regression)-based radial basis function neural network with dual extended Kalman filter," *Energy*, vol. 49, no. 1, pp. 413–422, 2013.

[245] Y. He and Y. Zheng, "Short-term power load probability density forecasting based on Yeo-Johnson transformation quantile regression and Gaussian kernel function," *Energy*, vol. 154, pp. 143–156, 2018.

[246] A. Kavousi-Fard, H. Samet, and F. Marzbani, "A new hybrid Modified Firefly Algorithm and Support Vector Regression model for accurate Short Term Load Forecasting," *Expert Syst. Appl.*, vol. 41, no. 13, pp. 6047–6056, 2014.

[247] X. Qiu, Y. Ren, P. N. Suganthan, and G. A. J. Amaratunga, "Empirical Mode Decomposition based ensemble deep learning for load demand time series forecasting," *Appl. Soft Comput. J.*, vol. 54, pp. 246–255, 2017.

[248] J. Bedi and D. Toshniwal, "Empirical Mode Decomposition Based Deep Learning for Electricity Demand Forecasting," *IEEE Access*, vol. 6, pp. 49144–49156, 2018.

[249] W. C. Hong and G. F. Fan, "Hybrid empirical mode decomposition with support vector regression model for short term load forecasting," *Energies*, vol. 12, no. 6, pp. 1–16, 2019.

[250] Y. Liang, D. Niu, and W. C. Hong, "Short term load forecasting based on feature extraction and improved general regression neural network model," *Energy*, vol. 166, pp. 653–663, 2019.

[251] J. Zhang, Y. M. Wei, D. Li, Z. Tan, and J. Zhou, "Short term electricity load forecasting using a hybrid model," *Energy*, vol. 158, pp. 774–781, 2018.

[252] M. R. Haq and Z. Ni, "A new hybrid model for short-term electricity load forecasting," *IEEE Access*, vol. 7, pp. 125413–125423, 2019.

[253] C. Li, X. Zheng, Z. Yang, and L. Kuang, "Predicting Short-Term Electricity Demand by Combining the Advantages of ARMA and XGBoost in Fog Computing Environment," *Wirel. Commun. Mob. Comput.*, vol. 2018, pp. 1–18, 2018.

[254] S. J. Huang and K. R. Shih, "Short-term load forecasting via ARMA model identification including non-Gaussian process considerations," *IEEE Trans. Power Syst.*, vol. 18, no. 2, pp. 673–679, 2003.

[255] R. Angamuthu Chinnathambi *et al.*, "A Multi-Stage Price Forecasting Model for Day-Ahead Electricity Markets," *Forecasting*, vol. 1, no. 1, pp. 26–46, 2018.

[256] Y. He, R. Liu, H. Li, S. Wang, and X. Lu, "Short-term power load probability density forecasting method using kernel-based support vector quantile regression and Copula theory," *Appl. Energy*, vol. 185, pp. 254–266, 2017.

[257] J. Che and J. Wang, "Short-term load forecasting using a kernel-based support vector regression combination model," *Appl. Energy*, vol. 132, pp. 602–609, 2014.

[258] J. Massana, C. Pous, L. Burgas, J. Melendez, and J. Colomer, "Short-term load forecasting in a non-residential building contrasting models and attributes," *Energy Build.*, vol. 92, pp. 322–330, 2015.

[259] M. S. Al-Musaylh, R. C. Deo, J. F. Adamowski, and Y. Li, "Short-term electricity demand forecasting with MARS, SVR and ARIMA models using aggregated demand data in Queensland, Australia," *Adv. Eng. Informatics*, vol. 35, pp. 1–16, 2018.

[260] J. W. Taylor, "Triple seasonal methods for short-term electricity demand forecasting," *Eur. J. Oper. Res.*, vol. 204, no. 1, pp. 139–152, 2010.

[261] P. Singh, P. Dwivedi, and V. Kant, "A hybrid method based on neural network and improved environmental adaptation method using Controlled Gaussian Mutation with real parameter for short-term load forecasting," *Energy*, vol. 174, pp. 460–477, 2019.

[262] K. Mason, J. Duggan, and E. Howley, "Forecasting energy demand, wind generation and carbon dioxide emissions in Ireland using evolutionary neural networks," *Energy*, vol. 155, pp. 705–720, 2018.

[263] L. Wang, S. X. Lv, and Y. R. Zeng, "Effective sparse adaboost method with ESN and FOA for industrial electricity consumption forecasting in China," *Energy*, vol. 155, pp. 1013–1031, 2018.

[264] W. Q. Li and L. Chang, "A combination model with variable weight optimization for short-term electrical load forecasting,"

*Energy*, vol. 164, pp. 575–593, 2018.
[265] J. Wu, Z. Cui, Y. Chen, D. Kong, and Y. G. Wang, "A new hybrid model to predict the electrical load in five states of Australia," *Energy*, vol. 166, pp. 598–609, 2019.
[266] Y. Hu *et al.*, "Short term electric load forecasting model and its verification for process industrial enterprises based on hybrid GA-PSO-BPNN algorithm—A case study of papermaking process," *Energy*, vol. 170, pp. 1215–1227, 2019.
[267] R. Wang, J. Wang, and Y. Xu, "A novel combined model based on hybrid optimization algorithm for electrical load forecasting," *Appl. Soft Comput. J.*, vol. 82, p. 105548, 2019.
[268] J. Wang, W. Yang, P. Du, and Y. Li, "Research and application of a hybrid forecasting framework based on multi-objective optimization for electrical power system," *Energy*, vol. 148, pp. 59–78, 2018.
[269] P. Singh and P. Dwivedi, "A novel hybrid model based on neural network and multi-objective optimization for effective load forecast," *Energy*, vol. 182, pp. 606–622, 2019.
[270] N. Zhang, Z. Li, X. Zou, and S. M. Quiring, "Comparison of three short-term load forecast models in Southern California," *Energy*, vol. 189, p. 116358, 2019.
[271] N. Ghadimi, A. Akbarimajd, H. Shayeghi, and O. Abedinia, "Two stage forecast engine with feature selection technique and improved meta-heuristic algorithm for electricity load forecasting," *Energy*, vol. 161, pp. 130–142, 2018.
[272] E. Paparoditis and T. Sapatinas, "Short-term load forecasting: The similar shape functional time-series predictor," *IEEE Trans. Power Syst.*, vol. 28, no. 4, pp. 3818–3825, 2013.

## Authors' information

[1] Department of Computer Engineering, University of Sharjah, Sharjah, UAE, 27272
Data Analytics and Cybersecurity Research Centre, RISE, University of Sharjah, Sharjah, UAE, 27272
[2] Department of Computer Engineering, University of Sharjah, Sharjah, UAE, 27272
Sustainable Energy & Power Systems Research Centre, RISE, University of Sharjah, Sharjah, UAE, 27272
[3] Department of Data Science, Princess Sumaya University for Technology, Amman 11941, Jordan
[4] Department of Electrical Engineering, University of Sharjah, Sharjah, UAE, 27272
[5] Distribution Control Centre, Dubai Water and Electricity Authority, Dubai, UAE

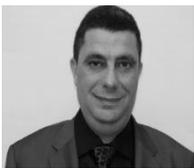

**Ali Bou Nassif** is currently an associate professor of Computer Engineering, as well as the Assistant Dean of Graduate Studies at the University of Sharjah, UAE. Ali is also an Adjunct Research Professor at Western University, Canada. Ali's research interests include software engineering, artificial intelligence, deep learning, natural language processing, speech processing, image processing, networking, security and E-Learning. Ali Has over 100 published conference and journal papers. Ali is a registered professional engineer (P.Eng) in Ontario, Canada.

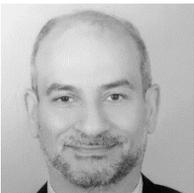

**Bassel Soudan** (S'89, M'94, SM'05) received the BS degree in electrical engineering from the EECS Department at the University of Illinois at Chicago in 1986, the MS and PhD degrees from the ECE Department at the Illinois Institute of Technology in 1988 and 1994.
From 1994 through 1996, he was with Design Technology group at Intel and from 1996 through 1999, he was with the Merced Microprocessor Design Team at Intel working on the development of the Itanium® processor. He has been with the ECE Department at the University of Sharjah since 1999. His current research interests involve the optimization of systems targeted towards renewable energy.

Dr. Soudan is a member of the IEEE, IEEE Computer Society, IEEE Circuits and Systems Society, ACM, and ACM SIGDA.

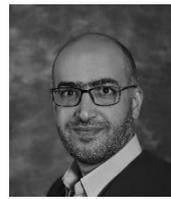

**Mohammed Azzeh** is a professor of Data Science at Princess Sumaya University for Technology. He holds a PhD in computing from University of Bradford, UK and MSc in Software Engineering from University of the West of England, UK. He is currently working as a faculty staff member in Data Science department. His research interests focus on Data Science, Machine Learning, Data mining, Empirical Software Engineering, Mining Software Repositories. Dr. Azzeh published over 40 research articles in reputable journals and conferences.

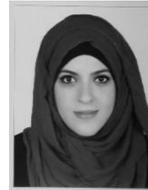

**Imtinan Atilli** received her B.Sc. degree in electrical and communications engineering from United Arab Emirates University, Al-Ain, UAE in 2009, and her M.Sc. degree in electrical and electronics engineering from the University of Sharjah, Sharjah, UAE in 2019. She also holds an M.Sc. in project management from the British University in Dubai, Dubai, UAE in 2013. She has been working as a Lab Engineer in the ECE department at UOS since 2010. Her research interests include mixed analog digital IC design, low voltage mixed mode CMOS circuits, operational transconductance amplifier circuit design, and operational amplifier circuit design. She is also interested in research related to the use of artificial intelligence tools in different applications such as speech and power load prediction.

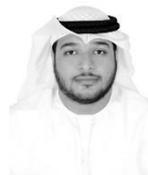

**Omar Almulla** is a senior Electrical Engineer in Dubai Electricity and Water Authority (DEWA). He works in Distribution Control Center where he monitors electrical loads (0.4, 6.6, 11 and 33 kilo Volts); He attends system emergencies for supply restoration and provide safe environment for attending any kind of electrical maintenance required. He holds B.S. in Electrical Engineering from the American University in Sharjah, UAE. Almulla has a work experience of 7 years in DEWA. He is a member if IEEE since 2013.

*A. B. Nassif, B. Soudan, M. Azzeh, I. Attilli, O. AlMulla*

# Appendix A – Table of Acronyms

| Acronym | Expanded Form | Acronym | Expanded Form |
|---------|---------------|---------|---------------|
| ACCE | Adaptive Circular Conditional Expectation | MAPE | Mean Absolute Percentage Error |
| ACF | Auto Correlation Function | MARA | multivariate auto regressive algorithm |
| ACM | Association of Computing Machinery | MARS | Multivariate Adaptive Regression Spline |
| ACO | Ant Colony Optimization | MAXPE | Maximum Percentage Error |
| AI | Artificial Intelligence | MFA | Modified Firefly Algorithm |
| AMPE | Average Modulus Percentage Error | ML | Machine Learning |
| ANFIS | Adaptive Neuro-Fuzzy Inference System | MLP | Multilayer Perceptron |
| ANN | Artificial Neural Network | MLR | Multiple Linear Regression |
| APE | Absolute Peak Error | MPC | Model Predictive Control |
| ARIMA | Autoregressive Integrated Moving Average | MPE | Mean Percentage Error |
| ARMA | Autoregressive-Moving Average | MRPE | Mean Regression Percentage Error |
| ARTV | Autoregressive Based Time Varying | MSE | Mean Squared Error |
| BGA | Binary Genetic Algorithm | MTLF | Medium-Time Load Forecasting |
| BNN | Bayesian Neural Networks | NARX | Nonlinear Autoregressive Exogenous |
| BooNN | Boosted NN | NCRPS | normalized continuous ranked probability score |
| BP | Back-Propagation | NMAE | Normalized Mean Absolute Error |
| BPNN | Back-Propagation Neural Network | NME | Normalized Mean Error |
| BVAR | Bayesian Vector Auto-Regression | NMSE | Normalized Mean Square Error |
| CNN | Convolutional Neural Network | NN | Neural Network |
| CVNN | Complex Valued NN | NRMSE | Normalized Root Mean Square Error |
| DME | Direct Measurable Error | PCA | Principal Component Analysis |
| DPE | Direct Percentage Error | PJM | Pennsylvania, Jersey, Maryland |
| DWT | Discrete Wavelet Transform | PMI | Partial Mutual Information |
| EBR | Entropy-Based Reduction | PSO | Particle Swarm Optimization |
| ELM | Extreme Learning Machine | RBF | Radial Basis Function |
| EMD | Empirical Mode Decomposition | RF | Radial Function |
| EMS | Energy Management System | RICNN | Recurrent Inception Convolution Neural Network |
| ER | Error Rate | RMAE | Root Mean Absolute Error |
| FASE | Forecast-Aided State Estimator | RME | Root Mean Error |
| FCM | Fuzzy C-Means | mRMR | minimal Redundancy Maximal Relevance |
| FOA | fruit fly optimization algorithm | RMSE | Root Mean Square Error |
| FTLRNN | Focused Time Lagged Recurrent Neural Network | RMSRE | root mean square ranging error |
| GA | Genetic Algorithms | RMSSE | Root Mean Square Scale Error |
| GEP | Gene Expression Programming | RNN | Recurrent Neural Network |
| GRNN | General Regression Neural Network | RSE | Relative Standard Error |
| IEEE | Institute of Electrical and Electronics Engineers | RVFL | Random Vector Functional Link network |
| LASSO | Least Absolute Shrinkage and Selection Operator | SARIMA | Seasonal ARIMA |
| LM | Levenberg–Marquardt | SARIMAX | Seasonal ARIMA with exogenous variables |
| LSSVM | Least Squares Support Vector Machine | SDAPE | Standard Deviations of Absolute Percentage Errors |
| LSTM | Long Short Term Memory | SDP | state dependent parameter |
| LTLF | Long-term load forecasting | SEP | Standard Error Prediction |

*A. B. Nassif, B. Soudan, M. Azzeh, I. Attilli, O. AlMulla*

| Acronym | Expanded Form | Acronym | Expanded Form |
|---|---|---|---|
| MAE | Median absolute error | SMBM | Smart Meter Based Models |
| SO | Simulation-Optimization | TDNN | Time Delay Neural Network |
| SOM | Self-Organizing Map Model | TSK | Takagi-Sugeno-Kang |
| SRWNN | Self-Recurrent Wavelet Neural Network | WGPR | Weighted Gaussian Process Regression |
| SSA | Improved Singular Spectral Analysis | WMAE | Weighted Mean Absolute Error |
| SSSC | Season Specific Similarity Concept | WMAPE | Weighted Mean Absolute Prediction Error |
| SSVRE | Subsampled Support Vector Regression Ensemble | WME | Weighted Mean Error |
| STLF | Short-term load forecasting | WNN | Wavelet Neural Network |
| SVD | Singular Value Decomposition | WRMSE | weighted root mean square error |
| SVM | Support Vector Machine | WVM | Weighted Voting Mechanism |
| SVR | Support Vector Regression | | |

*A. B. Nassif, B. Soudan, M. Azzeh, I. Attili, O. AlMulla*

# Appendix B – Detailed Data for each Study in the Review Pool

*VIII.1. Standalone Artificial Intelligence Models*

Table 1 – Classification of References from the Review Pool that use Standalone AI Models for STLF

| Classification | Reference | Algorithm | Evaluation Metrics |
|---|---|---|---|
| Artificial Neural Networks | [1] | Integrated ANN | - |
| | [14] | Cascaded NN | MAPE = 1.1273% |
| | [15] | Wavelet NN | WMAPE = 5.744% |
| | [16] | Input Vector (IV) based on ANN | MAPE = 1.38%, MPE = 0.1%, RSE = 0.63MW, and MSE = 0.40MW |
| | [17] | ANN | R2 = 0.9861, MSE = 4.89, MAPE = 5.56% |
| | [18] | ANN | - |
| | [19] | ANN | MAPE = 10.3%, and MAE = 3.63 kW |
| | [20] | ANN | MAPE = 3.09%, |
| | [21] | Hierarchical Neural Model | MAPE = 2.65% |
| | [22] | Wavelet Elman NN | MAPE = 2.29% |
| | [23] | Fractal Interpretation and Wavelet Analysis | Average relative error = 2.296% |
| | [24] | Wavelet Transform | Forecasting Error = 0.89% |
| | [25] | Wavelet NNs With Data Pre-Filtering | MAPE = between 0.08% to 0.49% |
| | [26] | Wavelet NN | $R^2$ = between 0.2 to 1 |
| | [27] | Wavelet De-Noising into a Combined Model | MAPE = 0.016, RMSE = 212.2451, and MAE = 164.5789 |
| | [28] | Wavelet NN | MAPE = 1.34% |
| | [29] | Radial Basis Function Algorithm | MAPE = 1.3% |
| | [30] | ANN | MAPE = 1.99% |
| | [31] | Multistage ANN | MAPE = 1.25% |
| | [32] | Generalization Learning Strategies | MAPE = 3.19% |
| | [33] | ANN | MAPE = 1.5% |
| | [34] | ANN | MAPE = 2.67% |
| | [35] | Gene Expression Programming (GEP) | R = 0.9462, MSE = 22664.0693, and MAE = 111.5629 |
| | [36] | Bayesian NN | RMSE = 4.28MW, and MAPE = 1.29% |
| | [37] | Bayesian Techniques, ANN | MAPE = 1.34% |
| | [38] | ANN | APE = 0.981 |
| | [39] | ANN | MAPE = 1.14%, and MAE < 0.0056 |
| | [40] | Boosted NN (BooNN) | MAPE = 1.42% |
| | [41] | ANN | RMSE = 3.68308% |
| | [42] | ANN | MAPE = 2.06%, and MAE = 15.27 MW |
| | [43] | ANN | MAPE = 0.66% |
| | [44] | ANN | Forecast error = 0.33% |
| | [45] | Bagged NNs | MAPE = 1.5% |
| | [46] | ANN | MAPE = 2% |
| | [47] | ANN | WME = 194%, VAR = 1.21% |
| | [48] | Integrated Algorithm for Optimizing Mapping Functions | APE = 1.833% |
| | [49] | ANN | NMAE = 7.65%, nRMSE = 17.9%, and WMAE = 33.6% |
| | [50] | Quantized Elman NN | MAPE = 1.444% |
| | [51] | ANN with a Clustering Algorithm | MAPE = 3.7% |
| | [52] | Back Propagation NN | - |
| | [53] | ANN | RME error = 1.4% |
| | [54] | ANNs using a Clustering Methodology | MAPE = 5.24% |
| | [55] | ANN | MAPE = 3.08% |
| | [56] | ANN | APE = 3.1% |
| | [57] | Novel Deep NN Architecture | MAPE = 1.405%, and MAE = 104.24 MW |
| | [58] | Extreme Learning Machine (ELM) Approach | MAPE = between 0.35% to 1.58%, and RMSE = between 54.9 to 71.92 |
| | [59] | Complex Valued NN (CVNN) | MAE = 57.17kW, and RMSE = 43.22 kW |
| | [60] | ANN | MAPE = 0.99%, and RMSE = 2.61% |
| | [61] | Smart Meter Based Models (SMBM) Tested with Well-Known Machine Learning Techniques ANN, SVM and Least Square SVM | – |
| | [62] | Deep NN Algorithm | MAPE = 9.77% |
| | [63] | Deep Feedforward Network | MAPE = 1%, MRPE = 3%, and MAE = 243 |
| | [64] | ANN | APE = 6.57% |
| Time Series Models | [66] | Stochastic Framework | - |
| | [67] | Soft Computing Based Technique | - |
| | [68] | A Constrained Orthogonal Least-Squares Method | APE = 1.65% |
| | [69] | Non-Linear Fractional Extrapolation Algorithm | MAPE = between 0.299% to 0.61% |
| | [70] | Improved Singular Spectral Analysis (SSA) Method | DME = 1.80%, DPE = 6.01% |
| | [71] | Multiple Equation Time Series | MAPE = 1.36% |
| | [72] | Generalized Long Memory (By Means of Gegenbauer Processes) | MAPE = 2.23% |
| | [73] | SARIMAX | MAPE = 2.98%, and RMSE = 62.61 MW |
| | | ANN | MAPE = 3.57%, and RMSE = 72.92 MW |

| Classification | Reference | Algorithm | Evaluation Metrics |
|---|---|---|---|
| | [74] | Time-Series Modeling with Peak Load Estimation capability | MAPE = 1.01% |
| | [272] | Novel Functional Time Series Methodology | RMAE = 0.0222, and MAE = 10.996 |
| | [76] | Periodic Time Series | MAPE < 1% |
| | [77] | Time Series-Oriented Load Prediction Model | - |
| | [78] | Vector Autoregressive | RMSSE = 0.3253 |
| | [79] | Autoregressive Based Time Varying (ARTV) | MAPE = 0.621%. |
| | [80] | Weighted Gaussian Process Regression (WGPR) Approach | nRMSE =6.614%, and nMSE = 0.020% |
| | [81] | Gaussian Processes | MAPE = 1.3%, NRMSE = 1.869%, NCRPS = 0.9927% |
| | [82] | Novel Information Theory-Based Approach | MAPE = 0.3% |
| Fuzzy Logic | [83] | Fuzzy Linear Regression Model | APE = 2% |
| | [84] | Fuzzy Logic | Percentage error is < ±3% |
| | [85] | Fuzzy Neural Collaboration | MAPE = 2.27% |
| | [86] | Neuro-Fuzzy Based Approach | MAE = 1.01%, RMSE = 1.21% |
| | [87] | Adaptive Neuro-Fuzzy Network | MSE = 0.0016 |
| | [88] | Fuzzy Logic Controller | Fuzzy error = between 1.2% and 3.2% |
| | [89] | Fuzzy C-Regression Method | APE = 1.67% |
| | [90] | Fuzzy Model | APE = 0.9% |
| | [91] | Fuzzy Neural Based on Chaotic Dynamics Reconstruction | MAPE = between 1.22% and 2.7% |
| | [92] | Fuzzy Logic Approach | Average absolute relative error = 3.9% |
| | [93] | Linguistics Fuzzy Time Series | MAPE = 1.1% |
| ARIMA | [95] | ARMA | MAPE = 1.87% |
| | [96] | ARIMA | MAPE = 2% |
| | [97] | ARMA | The average error = 1.6% |
| | [98] | ARIMA | MAPE = 2.19%, RMSE = 4.91%, and SEP = 2.65% |
| | [99] | ARIMA | MAPE = 0.87%, MAXPE= 1.44%, and SDAPE= 0.35% |
| | [100] | Multivariate time-series | MAPE = 0.52% |
| | [101] | ARIMA | MAPE = 1.2% |
| | [102] | ARIMA | RMSE = between 1.32% and 2.80% |
| | [103] | SARIMAX | MAPE = 0.7%, MAE = 21.7, and RMSE = 31.6 |
| Decomposition Models | [105] | Decomposition and Segmentation of the Load Time Series | MAPE = 0.0384% |
| | [106] | Singular Value Decomposition (SVD) | MAPE = 1.3% |
| | [107] | Seasonal Decomposition using Evolution for Parameter Tuning | MAPE = 1.7% |
| Grey Prediction | [109] | Grey System Theory | AMPE= 0.585% |
| | [110] | Grey Correlation Contest Modeling | Average Forecasting Error = 3.23% |
| | [111] | Improved Grey Forecasting Model | MAPE = 2.78% |
| Kalman Filtering | [113] | Kalman Filtering Algorithm | MAPE = between 0.13% to 0.58% |
| | [114] | Kalman Filtering Algorithm | M = 12.8%, P = 0.17% |
| PSO | [115] | PSO | APE = 1.11% |
| | [116] | PSO | MAPE = 2.42% |
| Self-Organizing Maps | [117] | Hierarchical SOM Model | MAPE = 2.75% |
| | [118] | Kohonen's SOM | MAPE = 2.18% |
| | [119] | Self-Organizing Fuzzy Neural Network | MAPE = 1.607%, and MAE = 44.56 MW |
| ACO | [120] | ACO | Absolute relative errors = 0.01% |
| GA | [121] | GA | MAPE = 1.9% |

### VIII.2. Standalone Statistical Models

Table 2 – Classification of References from the Review Pool that use Standalone Statistical Models for STLF

| Classification | Reference | Algorithm | Evaluation Metrics |
|---|---|---|---|
| Regression Models | [122] | Modified General Regression Neural Network | MAPE = 2.79% |
| | [126] | multiple linear regression | WRMSE score = 95.5 |
| | [127] | Statistical regression method | RMSE = 5%, and MAE = 4% |
| | [128] | recursive algorithm | MAPE = 2.68%, and APE = 2.81% |
| | [129] | periodic autoregressive model | MAPE = 3.42% |
| | [123] | Evolutionary Multi-objective Optimization of Kernel-Based | MAPE = 2.04% |
| | [124] | nonparametric functional methods: regression techniques | Seasonal error ER = 2.59% |
| | [125] | linear auto regressions | MAPE = 1.56% |
| | [131] | regression-based adaptive weather sensitive | MAE = 1.64%, and MALE = 2.73% |
| | [132] | Fuzzy Linear Regression Method | Average maximum percentage error = 3.57% |
| | [133] | Pattern-based local linear regression models | MAPE = 1.15% |
| | [134] | increment regression tree | MAPE = 2.8% |
| | [135] | Semi-Parametric Additive Model | MAPE = 2.14%, and MAE = 126.7 MW |
| | [136] | Random Forest Regressor, k-Nearest Neighbor Regressor and Linear Regressor | MAPE = between 1 and 2% |
| | [137] | Gaussian process quantile regression | MAPE = 1.73%, and RMSE = 3.79% |
| | [138] | clustering based load forecasting | MAPE = 2.28% |

| Classification | Reference | Algorithm | Evaluation Metrics |
|---|---|---|---|
| | [139] | functional vector autoregressive state space model | MAPE = 0.18% |
| | [140] | partially linear additive quantile regression | MAPE = 0.9697% |
| | [141] | least absolute shrinkage and selection operator (LASSO) | MAE = 4.9, and RMSE = 7.834 |
| | [142] | Part of the Holt-Winters model which also incorporates discrete-interval moving seasonalities | MAPE = 2.5% |
| | [153] | Regression Models | MAPE < 2% , and RMSE = 1.6% |
| | [154] | Regression Models | MAPE = between 8% to 9% |
| | [155] | Regression Models | MAPE = 0.9% |
| | [145] | SVR | RMSE = 0.76% |
| | [146] | SVR | MAPE = between 1.29% to 2.45% |
| | [147] | SVR | MAPE = 4.826%, and RMSE = 130.118 |
| | [148] | evolutionary SVM for regressive model | MSE = 20.77, MAE = 3.704, and MAPE = 2.307% |
| | [149] | SVM and ACO | RMSRE = 1.5% |
| | [150] | SVR | nRMSE = 1.91, and nMAE = 1.91 |
| | [151] | SVR | MAPE = 2.84%, and MASE = 0.55% |
| | [153] | SVR | MAPE < 2% , and RMSE = 1.6% |
| | [154] | SVR | MAPE = between 8% to 9% |
| SVR | [155] | SVR | MAPE = 0.9% |
| | [153] | BVAR | MAPE < 2% , and RMSE = 1.6% |
| BVAR | [154] | BVAR | MAPE = between 8% to 9% |
| | [155] | BVAR | MAPE = 0.9% |

*VIII.3. Hybrid Techniques*

Table 3 – Classification of References from the Review Pool that use Hybrid Models for STLF

| Reference | Algorithm | Evaluation Metrics |
|---|---|---|
| [156] | BNN learned by Hybrid Monte Carlo algorithm | MAPE = 0.607%, and RMSE = 11.4848 MW |
| [157] | RBF neural network with the adaptive neural fuzzy inference system (ANFIS) | RMSE= 0.0504 kW |
| [166] | Hybrid Wavelet Models | MAPE = 0.989% |
| [158] | ANN combined with an Analog Ensemble (AnEn) | RMSE = 7.07% |
| [159] | ANN combined with Principal Component Analysis (PCA) | MAPE = 0.01% |
| [160] | ANN combined with Bayesian Regularization Algorithm | $R^2$ = 0.902, and RMSE = 8.5% |
| [161] | ANN optimized through Hybrid GA-PSO | MAPE = 0.4% |
| [162] | ANN combined with sigmoid and MARA | MAPE = 1.24% |
| [163] | ANN combined with SVM | MAPE = 2.51%, and MAE = 2.69% |
| [164] | Bayesian NN, discrete wavelet transform, and GA | MAPE = between 0.185% to 0.41% |
| [165] | Recurrent Inception Convolution NN (RICNN) combining RNN and CNN | MAPE = 4.1% |
| [167] | Binary Genetic Algorithm (BGA) with Gaussian Process Regression and ANN | MAPE = 1.96%, |
| [168] | Convolutional NN (CNN) combined with Long Short Term Memory (LSTM) | MAPE = 2.4%, |
| [169] | CNN and Fuzzy Time Series | MAPE = 3.18%, RMSE = 1702.70, and MdRAE = 0.4537 |
| [170] | Focused Time Lagged Recurrent NN (FTLRNN) with Time Delay NN (TDNN) | APE = 1.67%, and MAPE = 1.82% |
| [171] | Fuzzy NN and Chaos Genetic Algorithm | MAPE = 1.26% |
| [172] | Grey-based approach and BPNN | MAPE = 3.1% |
| [173] | Hybrid Correction Method: combination of neural network and fuzzy logic. | MAPE = 1.43% |
| [174] | LSTM combined with CNN | MAPE = 1.45%, and RMSE = 1.83% |
| [175] | NN and Fuzzy Inference Method | Average percentage relative error = 1.78% |
| [176] | NN integrated using Genetic Algorithms optimized Back Propagation (GA-BP) | MAPE = 6.5% |
| [177] | RBF algorithm NN with the Adaptive Neural Fuzzy Inference System (ANFIS) | Average forecasting error = 1.669% |
| [178] | Recurrent NN (RNN) integrated with ANFIS | RMSE = 3.932, MAPE = 5.57%, MAE = 3.055, and $R^2$ = 0.821 |
| [179] | Self-Recurrent Wavelet NN (SRWNN) trained by Levenberg–Marquardt (LM) algorithm | nRMSE = 4.98%, and nMAE = 3.74% |
| [180] | Spiking combined with ANN | MAPE = 1.92% |
| [181] | Discrete Wavelet Transform (DWT), Empirical Mode Decomposition (EMD) and Random Vector Functional Link network (RVFL) | MAPE = 2.79%, and RMSE = 218.329 |
| [182] | Fuzzy logic and wavelet transform integrated with a generalized neural network | RMSE = 0.0486 kW |
| [183] | Neural-wavelet model and Bayesian method Automatic Relevance Determination | APE = 0.779% |
| [184] | wavelet echo state networks with neural reconstruction | MAPE = 1.7729%, and MAE = 0.0101 |
| [185] | Wavelet transform combined with an ANN | MAPE = 1.603%, and MAE = 3.4009 MW |
| [186] | Wavelet transform and evolutionary extreme learning machine | MAPE = 0.5554% |
| [187] | Wavelet transform and grey model improved by PSO algorithm | MAPE = 1.8% |
| [188] | Wavelet transform and neuro-evolutionary algorithm | MAPE = 0.99% |
| [189] | Wavelet transform and partial least squares regression | MAPE = 0.54% |
| [190] | Wavelet transform combined with Holt–Winters and weighted nearest neighbor models | MAPE = 0.96% |
| [191] | Wavelet transform and adaptive models | NMSE = 0.02314, and MAE = 2.01 |
| [192] | Wavelet combined with PSO, ANN, and Simulation-Optimization (SO) | MSE = 0.139 |
| [193] | Kohonen NN and wavelet transform | Mean percentage error = 0.78% |
| [194] | Two-dimensional wavelet based SDP models | MAPE = 1.9% |
| [195] | Embedding the discrete wavelet transform into NN | MAPE = 1.01% |
| [196] | Hybrid PSO–SVM | MAPE = 1.85% |

*A. B. Nassif, B. Soudan, M. Azzeh, I. Attilli, O. AlMulla*

| Reference | Algorithm | Evaluation Metrics |
|---|---|---|
| [197] | Evolutionary Algorithm (EA) and PSO | MAPE = 1.44%, and the diurnal load forecasting accuracy (A) = 98.36% |
| [198] | PSO and memetic algorithm | MAE = 61.15, and MAPE = 1.09% |
| [199] | PSO-SVR and Fuzzy C-Means (FCM) Clustering Techniques | Average Absolute Error = 1.218% |
| [200] | Kohonen's SOM as clustering method and RBF as classification method | Levene's Test for Equality of Variances and t-Test for Equality Means. No value listed. |
| [201] | SOM and SVM | MAPE = 2.42%, and MAE = 134.5 |
| [202] | SOM and clustering k-means algorithm | - |
| [203] | SOM and feed forward neural network | MAPE = 1.83%, and MAE = 71 |
| [204] | Fuzzy rule-based TSK model combined with a set of fuzzy regressions (FR) | APE = 1.69% |
| [205] | Cooperative ACO-GA | MAPE = 0.903036%, MAE = 1090.586, and RMSE = 1292.381 |
| [206] | Adaptive PSO based on gravitational search algorithm and the nonlinear multi-layer perceptron NN | MAPE = 1.82%, MAE = 146.3765, and RMSE = 206.2354 |
| [207] | Adaptive Circular Conditional Expectation (ACCE) combined with Adaptive Linear Model (LM) | MAE = 0.94% to 0.96%, and $R^2$ = 0.02% to 0.1% |
| [208] | A combination of Akaike and Bayesian methods | MAPE = 0.86%, RMSE = 1.24%, and MAE = 0.92% |
| [209] | Algorithm for congestion and other power system variables | MAPE = between 0.0525 to 0.0903, and RMSE between 5.026 to 7.05 |
| [210] | Average procedure, combined method, hybrid model and adaptive PSO algorithm | MAPE = 0.52% |
| [211] | Chaotic time series method combined with PSO | MAPE = 2.48% |
| [212] | Deviation analysis and an adaptive algorithm | MAPE = 4.66% |
| [213] | Echo State Networks and PCA Decomposition | NRMSE = 0.1164±0.0041 |
| [214] | Extreme Learning Machine with Kernel and PSO | MAPE = 5.6% |
| [215] | FA, SVM and Season Specific Similarity Concept (SSSC) | MAPE Five times lesser than the traditional methods. Value not mentioned. |
| [216] | Forecast-Aided State Estimator (FASE) and a Multilayer Perceptron (MLP) | NME = 1.39% |
| [217] | Fuzzy Inductive Reasoning and Evolutionary Algorithms | MAPE = 1.19% |
| [218] | Grey prediction with rolling mechanism | MAPE = 3.43% |
| [219] | Imperialist competitive algorithm with high order weighted fuzzy time series | MAPE = 0.83% |
| [220] | K-means followed by a Weighted Voting Mechanism (WVM) | Improvement in MAPE from 28% to 33.4%. |
| [221] | Knowledge Based System combining Affinity Propagation and Binary Firefly Algorithms | MAPE = between 1.23% to 3.35% |
| [222] | Least Absolute Shrinkage & Selection Operator (LASSO) and LSTM integrated temporal model | MAPE = 2.81%, and RMSE = 7.83 watt/m$^2$ |
| [223] | Least Squares Support Vector Machine (LSSVM) and AutoCorrelation Function (ACF) | MAPE = 0.7706%, MAE = 37.2227, and $R^2$ = 0.9934 |
| [224] | Energy Management System (EMS) based on a Model Predictive Control (MPC) strategy | Reduction by 14.1% in energy unbalance and of 8.7% in annual operational cost. |
| [225] | Multiple seasonal patterns and modified firefly algorithm | MAPE = 0.8799% |
| [226] | Neuro-evolutionary algorithm and chaotic feature selection | MAPE = 0.6024% |
| [227] | Non-symmetric Penalty Function, sum of squared error and fuzzy inference system | MAPE = 5.23% |
| [228] | Refined exponentially weighted fuzzy time series and an improved harmony search | MAPE = 1.4% |
| [229] | Robust functional principal component analysis and nonparametric models with functional both response and covariate | MAPE = 5.97% |
| [230] | Hybrid GOA-SVM model based on similar day approach | MAPE = 1.39% |
| [231] | Generalized Regression and a Probabilistic Neural Network | MAPE = 1.85% |
| [232] | Generalized Regression, a Neural Network, and fly optimization algorithm | minimum error value = 0.0018 |
| [233] | Nonlinear models (ANNs) with linear models (of the ARIMA type) | MAPE = 1.1% |
| [234] | Quantile regression neural network and triangle kernel function | MAPE = 1.59%, and MAE = 244.6189 MW |
| [235] | SARIMA and Holt-Winters-Taylor combined with Back-Propagation Neural Network (BPNN) and Adaptive Neuro-Fuzzy Inference System (ANFIS) | MAPE = 0.863% |
| [236] | Smooth transition autoregressive and neural network | MAPE = 1.07% |
| [237] | Stacking Ensemble of EA, ANN and RF followed by a Generalized Boosted Regression Model | sMAPE = 0.02, MAE = 513.50, and RMSE = 714.56 |
| [238] | SVR and Nonlinear Autoregressive Exogenous RNN (NARX RNN) | MSE = 131.98, and MAE = 9.37 |
| [239] | MLR combined with SVR and BPNN | MAPE = between 0.073 and 0.138 |
| [240] | Combined regression-fuzzy approach | MAE = 1.94, and RMSE = 2.41% |
| [241] | MLR and PSO | MAPE = 3.5% |
| [242] | Phase Space Reconstruction Algorithm and Bi-Square Kernel Regression | MAPE < 2.20%, RMSE < 30.0, and MAE < 2.30 |
| [243] | Seasonal exponential adjustment method and the regression model | MAPE = 4.48% |
| [244] | SVR-based RBF neural network with dual extended Kalman filter | MAPE = 0.6% |
| [245] | Yeo-Johnson transformation quantile regression and Gaussian kernel function | MAPE = 1.3979%, and MAE = 248.3677 MW |
| [246] | SVR and Modified Firefly Algorithm (MFA) | MAPE = 1.6909%, RMSE = 2.0604, and MAE = 22.5423 |
| [247] | EMD combined with Deep Learning | MAPE = 0.53%, and RMSE = 49.86 |
| [248] | EMD combined with Deep Learning | APE = between 5% to 8% |
| [249] | EMD combined with PSO and SVR | MAPE = 3.92%, RMSE = 142.41, and MAE = 9.04 |
| [250] | EMD, minimal Redundancy Maximal Relevance (mRMR), General Regression Neural Network (GRNN) and Fruit Fly Optimization Algorithm (FOA) | MAE = 9.5823, RMSE = 7.3550, MAPE = 0.8093% and TIC = 0.0052 |
| [251] | EMD, ARIMA, Wavelet Neural Network (WNN) optimized by FOA | MAPE varies from 0.76% to 0.92%, MAE = 67.51MW, and RMSE = 2.03 |
| [252] | EMD combined with T-Copula and deep-belief network | MAPE = 1.68%, and RMSE = 118.36 |

*A. B. Nassif, B. Soudan, M. Azzeh, I. Attilli, O. AlMulla*

| Reference | Algorithm | Evaluation Metrics |
|---|---|---|
| [253] | ARMA combined with XGBoost in a Fog Computing Environment | MAE score = 0.8318% |
| [254] | ARIMA Model and Non-Gaussian considerations | Relative Forecast Error = 1.37% |
| [255] | ARIMA+GLM | MAPE = 2.59% |
| | ARIMA+SVM | MAPE = 2.73% |
| | ARIMA+LOWESS | MAPE = 2.66% |
| | ARIMA+RF | MAPE = 3.12% |
| [256] | Kernel-based support vector quantile regression and Copula theory | MAPE = 0.81%, and MAE = 47.20 |
| [257] | Kernel-based support vector regression combination model | MAPE = 2.01%, and MAE score = 164.64 |
| [258] | Multiple Linear Regression (MLR), MLP and SVR | MAPE = 0.35% |
| [259] | MARS | MAE = 0.765%, and RMSE = 0.990% |
| | SVR | MAE = 0.945%, and RMSE = 1.211% |
| | ARIMA | MAE = 3.939%, and RMSE = 4.184% |
| [260] | Triple seasonal methods: double seasonal ARIMA, an adaptation of Holt–Winters exponential, and exponential smoothing method | MAPE = 1.5% |